\documentclass[a4paper,fleqn]{cas-dc}

\usepackage[numbers]{natbib}
\usepackage{newfloat}
\usepackage{listings}
\usepackage{amssymb}
\usepackage{amsmath}
\usepackage{makecell}
\usepackage{multirow}
\usepackage{booktabs}
\usepackage{graphicx}
\usepackage{subfigure}
\usepackage{bbding}
\usepackage[ruled,linesnumbered]{algorithm2e}
\def\tsc#1{\csdef{#1}{\textsc{\lowercase{#1}}\xspace}}
\tsc{WGM}
\tsc{QE}

\newtheorem{myDef}{Definition}

\newdefinition{rmk}{Remark}
\newproof{pf}{Proof}
\newproof{pot}{Proof of Theorem \ref{thm}}

\begin{document}
\let\WriteBookmarks\relax
\def\floatpagepagefraction{1}
\def\textpagefraction{.001}

\shorttitle{FairMigration}    

\shortauthors{YanMing Hu et al.}  

\title [mode = title]{Migrate Demographic Group For Fair Graph Neural Networks}  



%

\author[1]{YanMing Hu}

\fnmark[1]

\ead{huym27@mail2.sysu.edu.cn}


\credit{Conception of the study, Writing original
draft, Methodology, Software, Validation, Experiment, revision of the draft and Investigation}

\affiliation[1]{organization={School of Computer Science and Engineering},
            addressline={Sun Yat-sen University}, 
            city={GuangZhou},
            country={China}
            }

\affiliation[2]{organization={School of Software Engineering},
            addressline={Sun Yat-sen University}, 
            city={ZhuHai},
            country={China}
            }

%

\author[2]{TianChi Liao}
\fnmark[2]

\ead{liaotch@mail2.sysu.edu.cn}


\credit{Revision of the draft}


\author[1]{JiaLong Chen}
\fnmark[3]
\ead{chenjlong7@mail2.sysu.edu.cn}
\credit{Revision of the draft}

\author[1]{Jing Bian}
\fnmark[4]
\ead{mcsbj@mail.sysu.edu.cn}
\credit{Supervision and Funding acquisition}

\author[2]{ZiBin Zheng}
\fnmark[5]
\ead{zhzibin@mail.sysu.edu.cn}
\credit{Supervision and Funding acquisition}

\author[1]{Chuan Chen*}
\fnmark[6]
\ead{chenchuan@mail.sysu.edu.cn}
\credit{Supervision and Funding acquisition}

\cortext[1]{Corresponding author}



\begin{abstract}
    Graph Neural networks (GNNs) have been applied in many scenarios 
    due to the superior performance of graph learning. 
    However, fairness is always ignored when designing GNNs. 
    As a consequence, biased information in training data can easily affect vanilla GNNs, 
    causing biased results toward particular demographic groups 
    (divided by sensitive attributes, such as race and age). 
    There have been efforts to address the fairness issue.
    However, existing fair techniques generally divide the demographic groups by raw sensitive attributes
    and assume that are fixed.
    The biased information correlated with raw sensitive attributes will run through the training process
    regardless of the implemented fair techniques.
    It is urgent to resolve this problem for training fair GNNs.
    To tackle this problem, we propose a brand new framework, FairMigration, 
    which is able to migrate the demographic groups dynamically, 
    instead of keeping that fixed with raw sensitive attributes.
    FairMigration is composed of two training stages. In the first stage, 
    the GNNs are initially optimized by personalized self-supervised learning, 
    and the demographic groups are adjusted dynamically.
    In the second stage, the new demographic groups are frozen and 
    supervised learning is carried out under the constraints of new demographic groups and adversarial training.
    Extensive experiments reveal that FairMigration achieves a high trade-off 
    between model performance and fairness.
\end{abstract}




\maketitle

\section{Introduction}

\begin{figure*}[H]
    \centering
    \subfigure[Credit]{
        \includegraphics[scale=0.3]{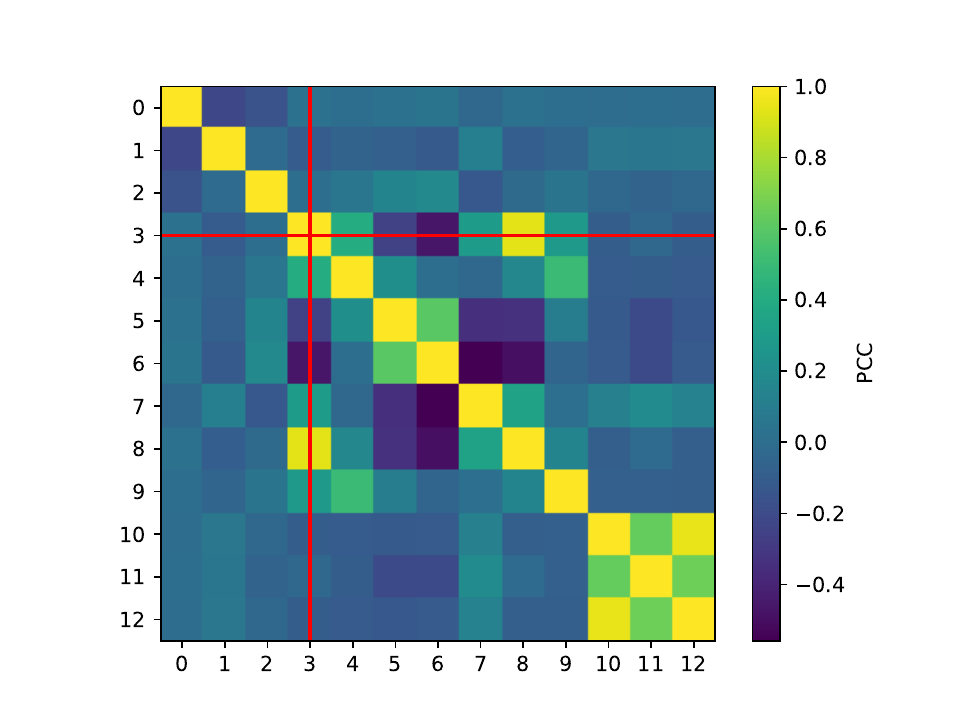}
        \label{credit_correlation}
    }
    \subfigure[Bail]{
        \includegraphics[scale=0.3]{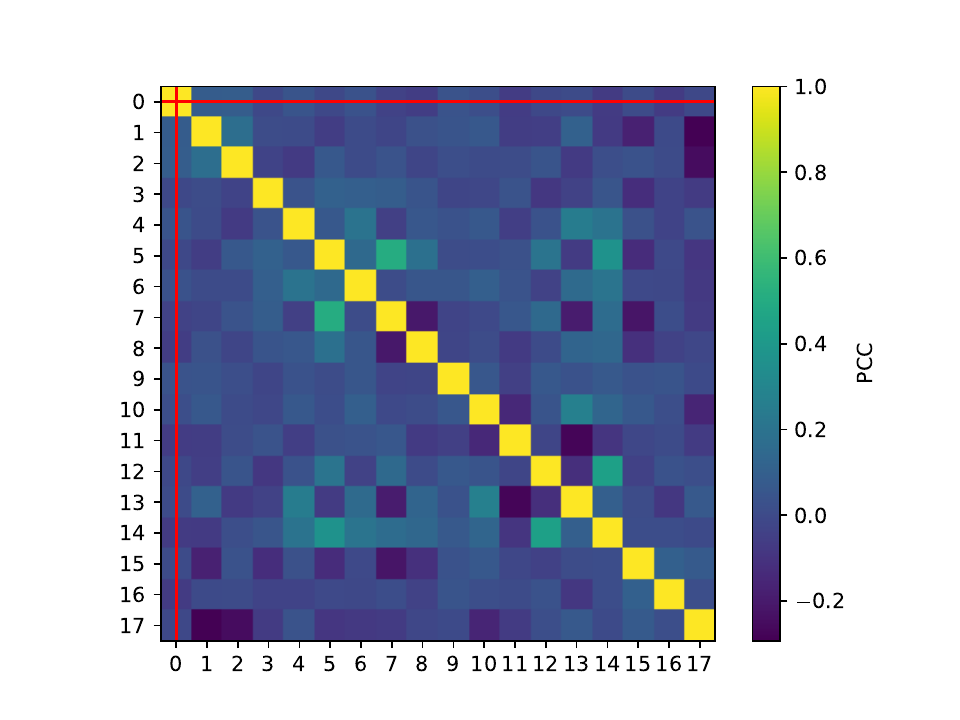}
        \label{bail_correlation}
    }
    \subfigure[Income]{
        \includegraphics[scale=0.3]{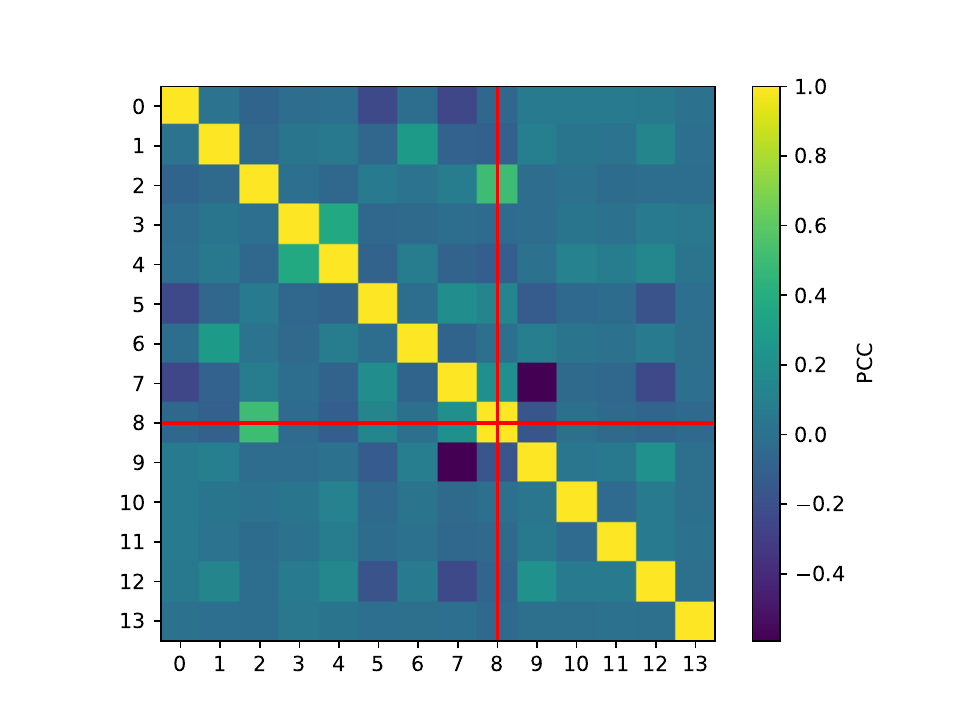}
        \label{income_correlation}
    }

    \caption{Visualization of attributes correlation. The cell in the i-th row and j-th column represents the Pearson Correlation Coefficient (PCC) among the i-th attribute and the j-th attribute. The sensitive attribute is marked by the red lines.}
    \label{vis_correlation}
\end{figure*}
In recent years, graph neural networks (GNNs) have attracted much attention due to the 
potent ability to convey data in a graph-structured manner. Many real-world scenarios, 
including node classification \cite{GCN,GAT,GIN,SAGE}, 
community detection \cite{com_dete_gnn_1,com_dete_gnn_2,com_dete_gnn_3}, 
link prediction \cite{link_1,link_2,link_3} 
and recommendation systems \cite{fair_recommend_redistribute,recom_1,recom_2,TAGCL}, 
have been applied to GNNs. 
The GNN aggregates the messages delivered by 
neighbors to obtain the embedding of nodes, edges, or graphs. 
The key to the powerful expression ability of GNN lies in that the GNN is able to
extract the attribute features and the structural features 
in the graph structure data at the same time.
However, the fairness issue in GNNs attracts very little attention. 

Recently, fairness has become a high-profile issue.
A number of recent studies have shown that AI models may produce biased results 
for specific groups or individuals divided by some 
particular sensitive attributes, such as race, age, gender, etc. 
The biased algorithms have resulted in some horrible cases.
An American judicial system's algorithm erroneously predicts that 
African Americans will commit twice as many crimes as whites. 
Amazon found that its recruitment system was biased against female candidates, 
especially in tech positions \cite{mehrabi2021survey}.
The further application of AI models will be severely constrained
if the fairness issue cannot be satisfactorily addressed.

In deep learning, there have been some works that try to resolve the fairness issue \cite{fair_boosting,counterfactual_text_cls,wang2023toward,Jalal2021fairness,fair_oversampling}. 
However, fair deep learning methods 
seldom take samples' interaction into account and the nodes in the graph interact with each other frequently.
Therefore, the fair deep learning techniques are not applicable to graph data \cite{graph_survey}. There are also several efforts to address the fairness issue in GNNs.
The works on fair graph representation learning adopt techniques like adversarial training \cite{FAIRGNN,fairac,fair_graph_rec}, self-supervised learning \cite{nifty,GEAR,GRADE,kose2021fairness,fairaug}, and limiting the distance of group distribution \cite{SIND,Edits,fairGAE}. These algorithms are still restricted to fixed sensitive information.  

The biased results generated by vanilla GNN can be attributed to a variety of factors.
\textbf{1) Subjective factors and historical impacts.} When labeling the data, 
it will ineluctably be influenced by subjective elements and 
resemble previous historical data \cite{his_fair_1,his_fair_2,his_fair_3,FAIRGNN}. 
These biased properties are introduced into graph neural 
networks during training and displayed in the outcomes. {{However, it is almost impossible to remove the influence of subjective factors in training data completely.}}
\textbf{2) Feature correlation.} GNNs correlate each dimension feature, 
including sensitive attributes, with the prediction result \cite{adagnn,fairVGNN,fairaug}. 
For different groups divided by sensitive attributes, 
the tightness and accuracy of the correlation 
between sensitive attributes and predicted labels may differ enormously.
{The Figure \ref{vis_correlation} demonstrates the correlations between different attributes on three datasets by the Pearson Correlation Coefficient (PCC). It can be seen that there are varying degrees of correlation between sensitive and non-sensitive attributes, as well as between non-sensitive and non-sensitive attributes.} 
\begin{figure*}[t]
    \centering
    \includegraphics[scale=0.15]{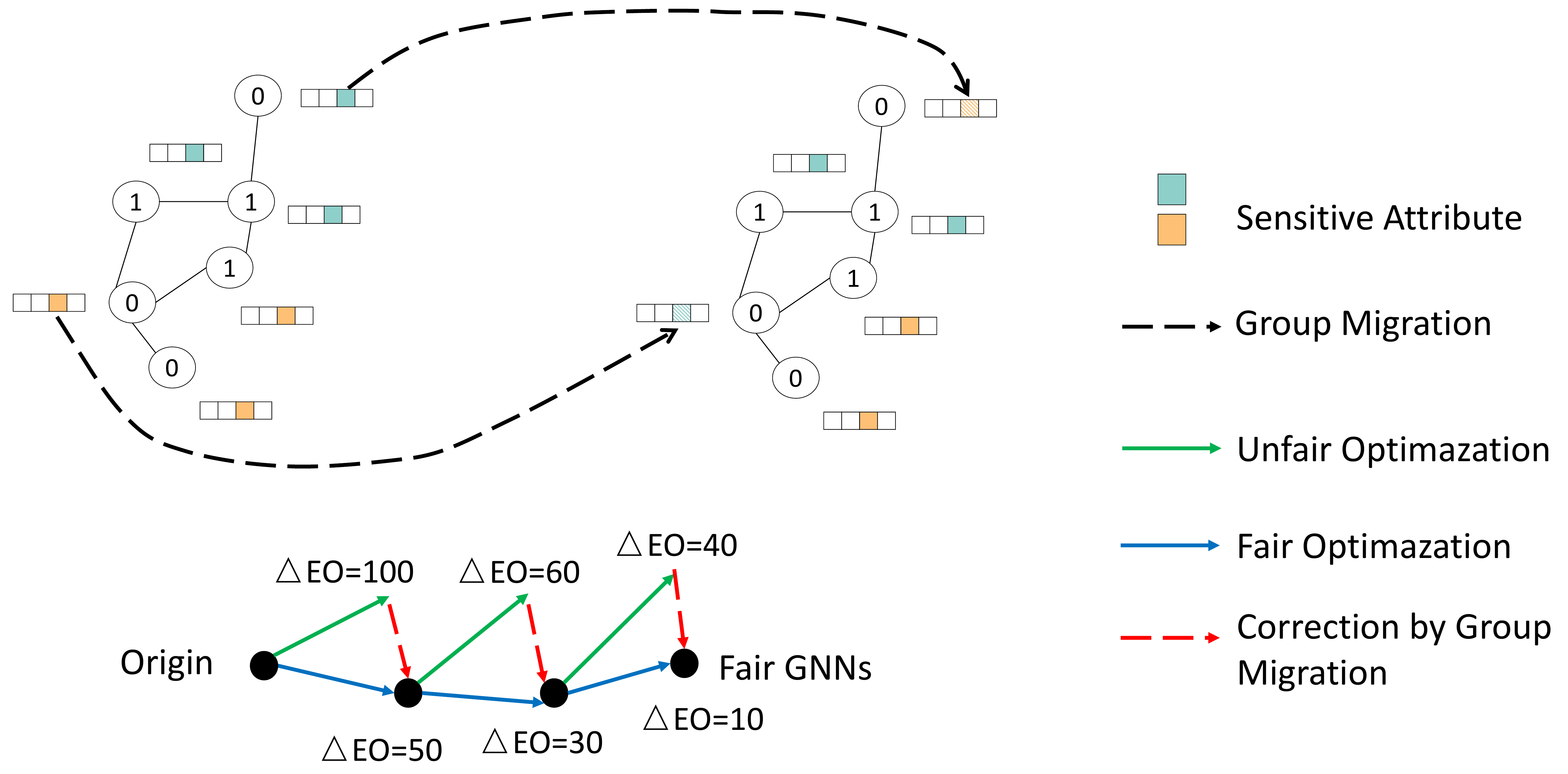}
    \caption{A toy diagram of group migration. $\Delta EO$ is a fairness metric. The lower $\Delta EO$ indicates the more fair performance. }
    \label{fig:mig_diag}
\end{figure*}
\begin{figure}[b]
    \centering
    \subfigure[Mean of distribution]{
        \includegraphics[scale=0.18]{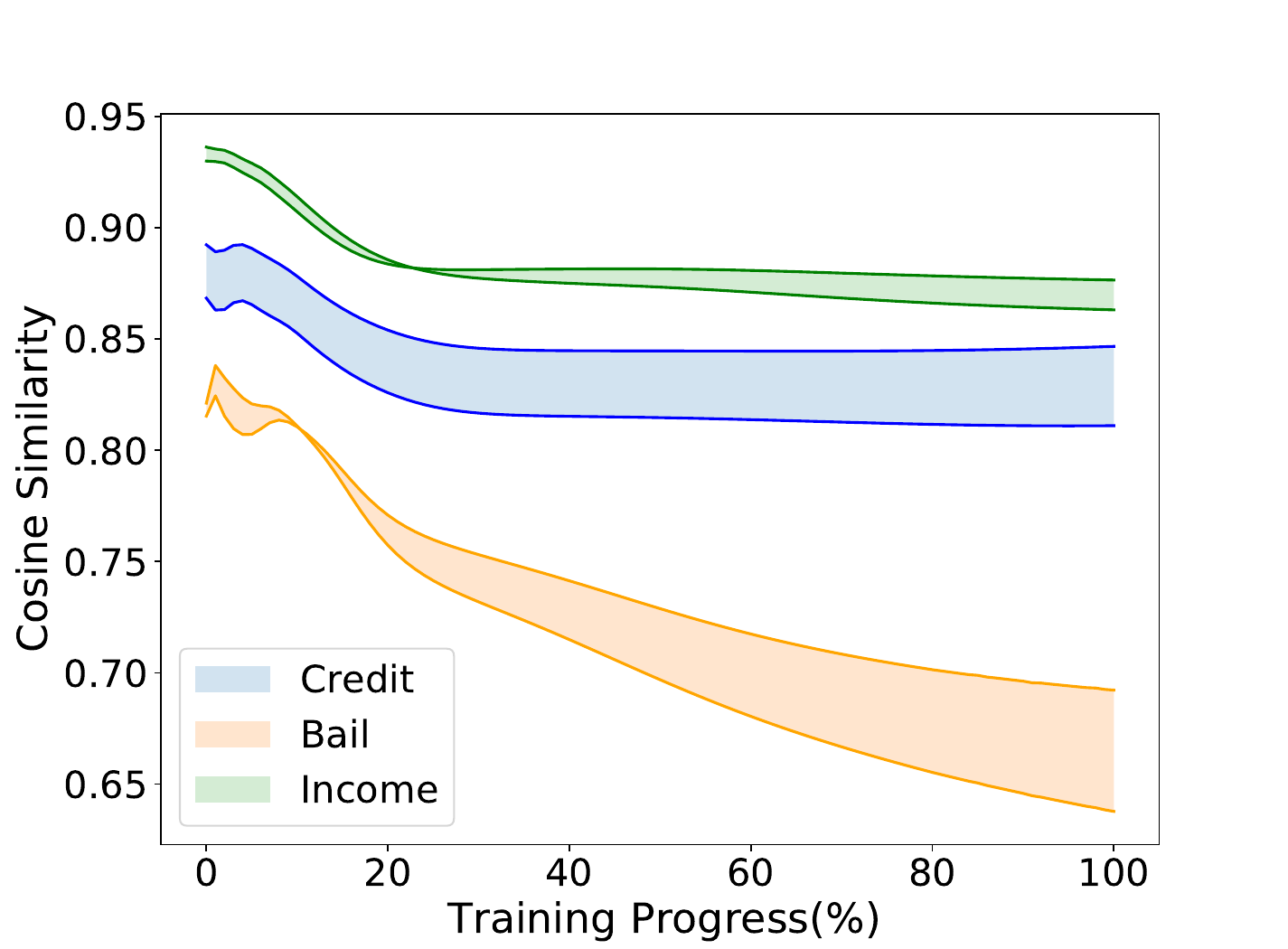}
        \label{gcn_vanilla_mean}
    }
    \subfigure[Std of distribution]{
        \includegraphics[scale=0.18]{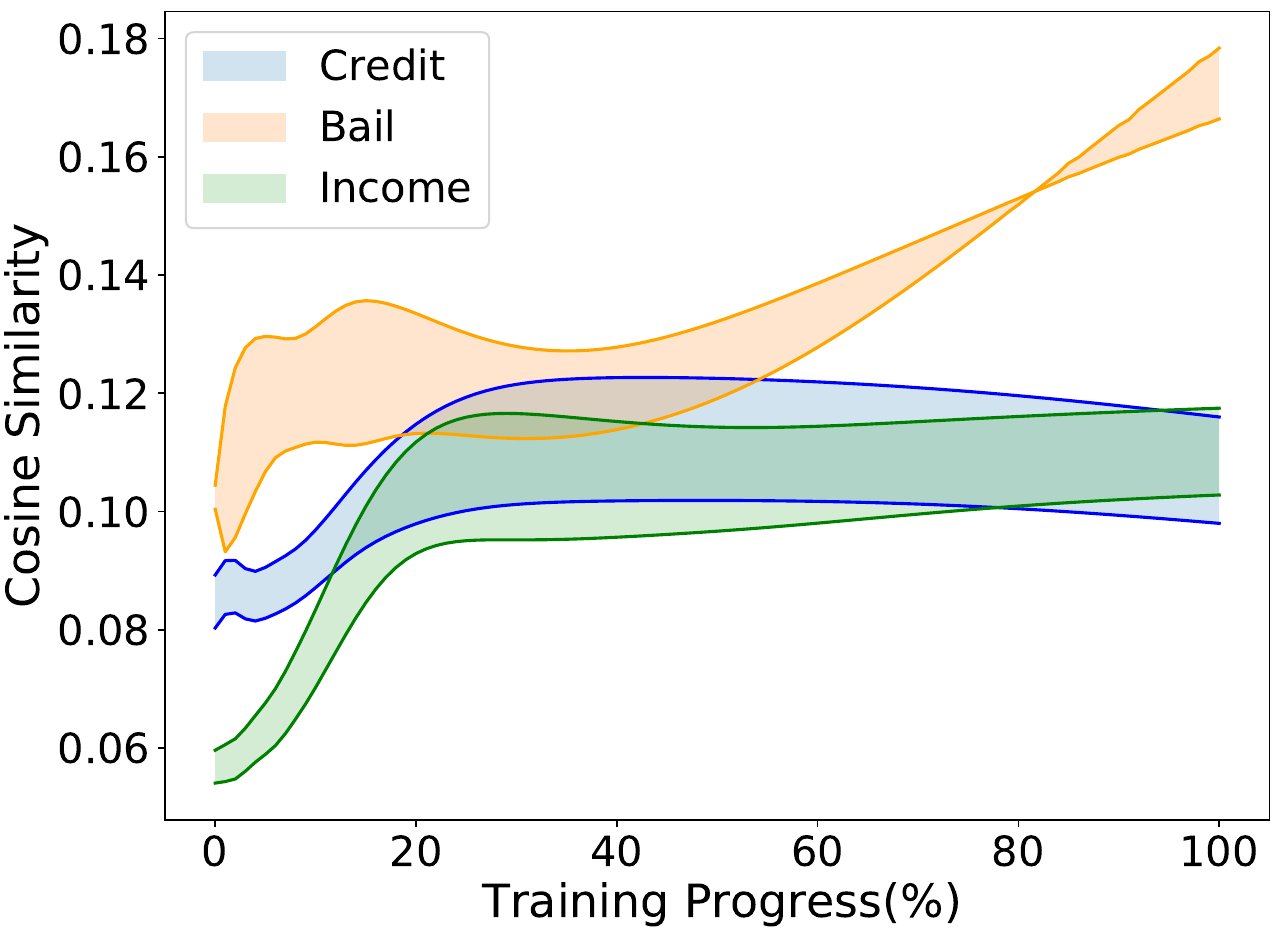}
        \label{gcn_vanilla_bail_mean}
    }
    \caption{The change curve of group similarity distribution of vanilla GCN on three datasets. 
    The bold lines with the same color indicate the mean / std of two 
    group similarity distributions on the same dataset, respectively.
    The shaded areas indicate the group similarity distribution difference. 
    The wider the shaded area on the y-axis, the larger the similarity difference between groups.}
    \label{trend}
\end{figure}
Even though removing the sensitive attributes, there are still some attributes that imply the sensitive attributes \cite{GEAR,Edits}. For example, height and weight imply gender.
Meanwhile, simply removing the sensitive attributes unavoidably causes the loss of useful information \cite{fpgnn}.
{At present, there is a lack of effective enough methods to guide graph neural networks to establish a reasonable causal relationship between the attributes and the labels.}
\textbf{3) Message propagation.} The topology of the graph also introduces biased information during message propagation. 
{In real-life scenarios, users are more likely to establish connections with similar individuals or download preferred apps. Meanwhile, the biased recommendations related to sensitive attributes dominate the information flow. Consequently, mapping this relationship to graph data, the number of intra-group edges greatly surpasses the number of inter-group edges. The bias in the topology structure is further amplified during the message propagation, and different groups will be more opposed and exclusive \cite{FMP,GCN,DEA,fairadj,fairdrop,dong2023fairness_survey,zhang2022trustworthy_survey}. }
Worse still, message propagation even changes the correlation of the non-sensitive attributes with the sensitive attributes and then further introduces bias \cite{fairVGNN}.
{The existing message-propagation-based methods are mostly designed for specific graph neural networks or require prior knowledge to preprocess training data, which is inconvenient for application and expansion.}
\textbf{4) Fixed sensitive attributes.} Sensitive attributes are considered immutable by existing fair algorithms on GNNs. 
Under such circumstances, the bias in the initial data will always 
impede the model training process through feature correlation and message propagation, regardless of how the model is improved. {Worse still, the influence of sensitive attributes may fluctuate during training. It is difficult to divide groups based on sensitive attributes in advance. }

Since training GNNs with fixed sensitive attributes causes unfair predictions and revising the sensitive attributes dynamically for fairness is unexplored, we might as well modify the sensitive attributes dynamically { (we call this operation group migration) to offset the biased influence. 
{The subjective factors and historical impacts can be reflected in the sensitive attributes. For example, in a prediction task, the prediction target is credit information and the sensitive attribute is age. The prediction results should be consistent instead of changing with the sensitive attributes. The solution can be reducing the performance gap between groups divided by the sensitive attributes.}
Through group migration, the direction and degree of feature correlation can be adjusted. The incorrect causal relationships between features would be decreased. Additionally, the group distribution of nodes and edges in the graph can be reconstructed, preventing biased message propagation. Moreover, the limitation of fixed sensitive attributes is broken, and the unfair influence can be alleviated. Group migration can be further expanded in future work, not only applicable to groups divided by sensitive attributes.}
Take an example with binary groups. Flipping the sensitive attributes of some nodes would change the optimization direction against the unfair results, which is shown in Figure \ref{fig:mig_diag}.
When training vanilla GCN \cite{GCN} on the three datasets, 
the change tendency of group similarities distribution (the definition can be seen in Definition 1.) is illustrated in Figure \ref{trend}. 
It is evident that the group similarity distribution between the two groups diverges 
increasingly with fixed sensitive attributes, 
leading to a biased prediction towards one group.

In this paper, we propose a novel model, FairMigration, for the 
fairness issue of GNNs. 
FairMigration is an additive method realizing the group migration, 
which could break the limitation of static sensitive attributes and 
gradually remove the bias 
information contained in the original data in the training of the GNNs.
The training of FairMigration is divided into two stages. 
In the first stage, the encoder is trained by personalized self-supervised learning and
learns the embeddings of nodes.
Based on group similarity distribution, outliers in one group are transferred to another group. 
After the first stage of training, 
the encoder is preliminarily trained and the migrated groups division is acquired. 
In the second stage, the encoder and the classifier are optimized by supervised learning 
under the condition of the new pseudo-demographic groups and adversarial training. 
The purpose of adopting a two-stage training framework is to prevent biased prediction from destroying group migration. 
Our contributions are summarized below:
\begin{itemize}
    \item To our knowledge, 
    we first point out that the biased information contained 
    in static sensitive attributes will always exist in 
    training for fair GNNs.
    \item We propose a model with group migration to 
    address the problem of biased information contained in static sensitive attributes.
    \item Extensive experiments verify the effectiveness of 
    the proposed method in this paper.
\end{itemize}

\section{Related Work}
\subsection{Graph Neural Networks}
Many Graph Neural Networks (GNNs) have been proposed to learn the 
representation of graph structure data.
GCN \cite{GCN} uses the first-order approximation of Chebyshev's 
polynomial as the aggregation function of message passing, 
GAT \cite{GAT} uses the attention mechanism to assign message weight for aggregation, 
GIN \cite{GIN} aggregates messages with a summation function.
Jump Knowledge (JK) \cite{jk} passes the output of each layer of the graph neural network to 
the final layer for aggregation.
APPNP \cite{APPNP} combines personalized PageRank with GCN to
aggregate the information of high-order neighbors. 
GraphSAGE \cite{SAGE} generalizes GCN to inductive tasks, learning a function that aggregates 
the representation of known neighbors.

\subsection{Fairness In Deep Learning}
The methods of solving the fairness problem in DL can be divided into three types: 
pre-processing, in-processing treatment, and post-processing \cite{mehrabi2021survey}.

The pre-processing methods perform debiasing operations on the data before training, 
such as modifying attributes and regenerating labels. 
Lahoti et al. \cite{lahoti2019operationalizing} create fair representation
before training through external knowledge.
Ustun et al. \cite{USTUN2019fairness} construct a debiased classifier through recursive feature picking.
Mehrotra et al. \cite{mehrotra2021mitigating} denoise sensitive attributes to reduce 
gender and racial bias.

The in-processing methods alleviate the bias of the model when training.
Chai et al. \cite{fair_adaptive_weights} dynamically adjust the weights of the loss function during training.
FairSmooth \cite{fair_gaussian_smooth} trains classifiers in each group separately 
and then aggregates classifiers by Gaussian smooth.
FairRF \cite{fairRF} explores the issue of training fair classifiers under the 
condition of unknown sensitive attributes.

The post-processing methods modify the output to obtain debiased results.
Lohia et al. \cite{lohia2019bias} use an individual bias detector for prioritizing data 
samples in a bias mitigation algorithm.
Mishler et al. \cite{mishler2021fairness} have developed a post-processing predictor that estimates, 
expands and adjusts previous post-processing methods through a dual robust estimator.
Putzel et al. \cite{putzel2022blackbox} modify the predictions of the 
black-box machine learning classifier for fairness in multi-class settings.

\subsection{Fairness In Graph Neural Networks}
There have been some attempts to address the group-level fairness issues of graph neural networks.
{In this paper, we study the fair GNN in a supervised learning manner. Most existing methods adopt pre-processing or in-processing and are commonly proposed for link prediction tasks or node classification tasks. Some representative methods are listed as follows.
FairAdj \cite{fairadj} is an in-processing method for link prediction. It exploits graph variational autoencoder and two different optimization processes to adjust the edge weights for a fair adjacency matrix and reduce performance differences between intra-group links and inter-group links.
FairDrop \cite{fairdrop} is an in-processing method for link prediction. It adds heterogeneous edges to the origin adjacency matrix and randomly drops some edges to break the filter bubble in link prediction.
FairGNN \cite{FAIRGNN} is an in-processing method for node classification. It uses adversarial training to prevent the leakage of sensitive attributes.
NIFTY \cite{nifty} is an in-processing method for node classification. It adopts a counterfactual augmented siamese network and regularization of the model 
parameters meeting the Lipschitz condition for debiasing.
Edits \cite{Edits}  is a pre-processing method for node classification. It constrains the Wasserstein distance of attributes between 
groups, and then augments topology based on preprocessed attributes to output debiased 
results. 
SIND \cite{SIND} is an in-processing method for node classification. It explores the influence of nodes on biased results and removes the biased nodes to align the group fairness.}
The above algorithms, failing to break the bias from the original demographic group division, train fair GNN under the fixed sensitive attributes and thus inevitably  
encounter bottlenecks of fairness improvement.
The connections and differences between FairMigration and the existing works are summarized in Table \ref{summary}. {Our work mainly concentrates on node classification. Thus, we select FairGNN, NIFTY, Edits, and SIND for experiment comparison.}
\begin{table}[htbp]
    \centering
    \caption{The summary of FairMigration and related works. AD denotes adversarial training used. SSL denotes self-supervised learning used. GD denotes the group distribution used. GM denotes group migration used. {In denotes in-processing. Pre denotes pre-processing. LP denotes link prediction. NC denotes node classification.}}
        \begin{tabular}{l|cccccc}
        \toprule
        Methods & Type&{Task} & AD & SSL & GD & GM \\
        \midrule
        {FairAdj}& {In} &{LP} & {\XSolidBrush} & {\XSolidBrush} & {\Checkmark} & {\XSolidBrush} \\        
        {FairDrop}& {In} &{LP} & {\XSolidBrush} & {\XSolidBrush} & {\Checkmark}& {\XSolidBrush} \\
        
        \midrule
        FairGNN& In  &{NC} & \Checkmark& \XSolidBrush & \XSolidBrush& \XSolidBrush\\
        NIFTY& In &{NC} & \XSolidBrush & \Checkmark & \XSolidBrush & \XSolidBrush\\
        Edits& Pre &{NC} & \XSolidBrush & \XSolidBrush& \Checkmark & \XSolidBrush \\
        SIND& In &{NC} & \XSolidBrush & \XSolidBrush & \Checkmark& \XSolidBrush \\
        \midrule
        FairMigrations& In &{NC}& \Checkmark & \Checkmark & \Checkmark & \Checkmark\\
        \bottomrule
        \end{tabular}%
    \label{summary}%
\end{table}%

\section{Preliminary}
In this section, the notations of this paper will be illustrated 
and the problem definition will be given.
\subsection{Notations}
Given a graph $\mathcal{G} (\mathbf{X}, \mathbf{A})$, $\mathbf{X} \in \mathbb{R}^{N \times K}$ 
denotes the 
attribute matrix of $\mathcal{G}$ and $\mathbf{A} \in \mathbb{R}^{N \times N}$ 
denotes the adjacency matrix of
$\mathcal{G}$. The sensitive attribute vector is denoted by $\mathbf{S} \in \mathbb{R}^N$, 
which is a column of the attribute matrix $\mathbf{X}$. 
$\mathbf{U} \in \mathbb{R}^{N \times {(K-1)}}$ 
denotes the attribute matrix removed $\mathbf{S}$ from $\mathbf{X}$.
$\mathcal{G} (\mathbf{X}, \mathbf{A})$ can be augmented to the j-th view 
$\tilde{\mathcal{G}}_j (\tilde{\mathbf{X}}_j, \mathbf{A})$, 
where $\tilde{\mathbf{X}}_j$ is the corresponding augmented attribute matrix.
GNN learns the representation of nodes, mapping the graph $\mathcal{G}$ into the embedding matrix
$\mathbf{Z} \in \mathbb{R}^{N \times d}$.
Similarly, the embedding matrix of graph $\tilde{\mathcal{G}}_j$ 
is denoted by $\mathbf{Z}_j\in \mathbb{R}^{N \times d}$.
The decoder GNN recovers the attribute matrix from $\mathbf{Z}$. 
The reconstructed attribute matrix is denoted by $\mathbf{X}^{rec}\in \mathbb{R}^{N \times K}$.
The reconstructed sensitive vector is denoted by $\mathbf{S}^{rec}\in \mathbb{R}^N$.
The matrix, removing $\mathbf{S}^{rec}$ from $\mathbf{X}^{rec}$, 
is denoted by $\mathbf{U} \in \mathbb{R}^{N \times {(K-1)}}$.
Our method involves a group migration module. The corresponding notation mainly includes the
current pseudo attribute matrix $\mathbf{P}\in \mathbb{R}^N$, 
the group similarity distribution matrix $\mathbf{Q}\in \mathbb{R}^N$, 
the prototype of i-th group $\mathbf{T}_i \in \mathbb{R}^{ d}$ and the outlier set
$\mathbf{O}$. For convenience, all the important notations are listed in Table \ref{tab:notation}.

\begin{table}[t]
    \centering
    \caption{Notation Statements Table}
        \begin{tabular}{l|l}
        \toprule
        \multicolumn{1}{l|}{Notation} & Statements \\
        \midrule
        \multicolumn{1}{l|}{$\mathcal{G} (\mathbf{X}, \mathbf{A})$} & original graph \\
        $\mathbf{X}$& attribute matrix\\
        $\mathbf{A}$& adjacency matrix\\
        $\mathbf{Y}$& ground-truth labels\\
        $\tilde{\mathbf{Y}}$& predicted labels\\
        $\mathbf{S} $ & sensitive attribute vector \\
        $\mathbf{U} $ & attribute matrix without sensitive attribute \\
        $\tilde{\mathcal{G}}_j (\tilde{\mathbf{X}}_j, \mathbf{A})$& j-th augmented graph \\
        $\tilde{\mathbf{X}}_j$ & j-th augmented attribute matrix \\
        $\mathbf{Z}$ & embedding of graph $\mathcal{G}$\\
        $\mathbf{Z}[i,:]$ & i-th row of  $\mathbf{Z}$\\
        $\mathbf{Z}_j$ & embedding of j-th augmented graph \\
        $\mathbf{Z}_j[i,:]$ & i-th row of $\mathbf{Z}_j$ \\
        $\mathbf{X}^{rec}$ & reconstructed attribute matrix \\
        $\mathbf{U}^{rec}$ & \makecell[l]{reconstructed attribute matrix \\without sensitive attribute} \\
        $\mathbf{S}^{rec}$& reconstructed sensitive attribute vector \\
        $\mathbf{S}_i^{rec}$& reconstructed sensitive attribute of i-th node \\
        $\mathbf{P}$ & current pseudo attribute matrix \\
        $\mathbf{T}_i$ & prototype of i-th group \\
        $\mathbf{Q}$  & group similarity distribution matrix \\
        $\mu_{\mathbf{P}_i}, \sigma_{\mathbf{P}_i}$ & mean and std of similarity of group ${\mathbf{P}_i}$ \\
        $\mathbf{O}$  & outlier set \\
        $\alpha, \gamma, \delta, \lambda$& hyperparameter \\
        \bottomrule
        \end{tabular}%
    \label{tab:notation}%
  \end{table}%
\subsection{Problem Definition}
The fairness issue can be divided into two levels: 
group-level fairness and individual-level fairness.
This paper mainly focuses on \textbf{single-dimension binary group-level} fairness.
The group-level fairness emphasizes fairness between different groups 
and treats every group equally.
The optimization of the fairness issue and model utility may differ vastly. For example, random prediction treats every sample equally, but the model utility is unacceptable.
The target of the fairness issue is to reduce the performance difference across groups while avoiding significant utility deterioration.

\begin{myDef}\textbf{Group Similarity distribution.}
Given a graph $\mathcal{G} (\mathbf{X}, \mathbf{A})$ and the corresponding pseudo sensitive attribute vector $\mathbf{P}$. The nodes in $\mathcal{G}$ can be divided into several groups according to $\mathbf{S}$. Node $v_i$, whose sensitive attribute is j, belongs to the j-th demographic group. The prototype of the j-th demographic group $\mathbf{T}_j$ can be defined as:
\begin{equation}
    \mathbf{T}_{j} = {\frac{1}{|\mathbf{P}_i = {j}|}}\sum_{i \in \{\mathbf{P}_i = {j}\}} \mathbf{Z}_i,
    \label{centerj}
\end{equation}
where $\mathbf{Z}_i$ is the embedding of node $v_i$.
The group similarity distribution of the j-th demographic group can be described by the mean value $\mu_j$ and standard deviation $\sigma_j$ of the cosine similarity set $\{cos(\mathbf{Z}_i, \mathbf{T}_{j})\quad|\quad\mathbf{P}_i = {j} \}$:
\begin{equation}
    \left\{
        \begin{aligned}
            & \mu_j = mean(\{cos(\mathbf{Z}_i, \mathbf{T}_{j})\quad|\quad\mathbf{P}_i = {j} \}), \\
    &\sigma_j = std(\{cos(\mathbf{Z}_i, \mathbf{T}_{j})\quad|\quad\mathbf{P}_i = {j} \}).
        \end{aligned}
    \right.
    \label{sup_loss}
\end{equation}
\end{myDef}

\begin{myDef}\textbf{Group Migration.}
    Given a node $v_i$, if $v_i$ deviates from the current 
pseudo demographic group significantly, it will be transferred to other demographic groups by revising the sensitive attribute.
Given a sensitive attribute modification function $\rho(\cdot)$, the group migration can be defined as:
\begin{equation}
    {\mathbf{P}_i^{new} = \rho(\mathbf{P}_i^{old}).}
\end{equation}
\end{myDef}



\section{Method}
\begin{figure*}
    \centering
    \includegraphics[scale=0.2]{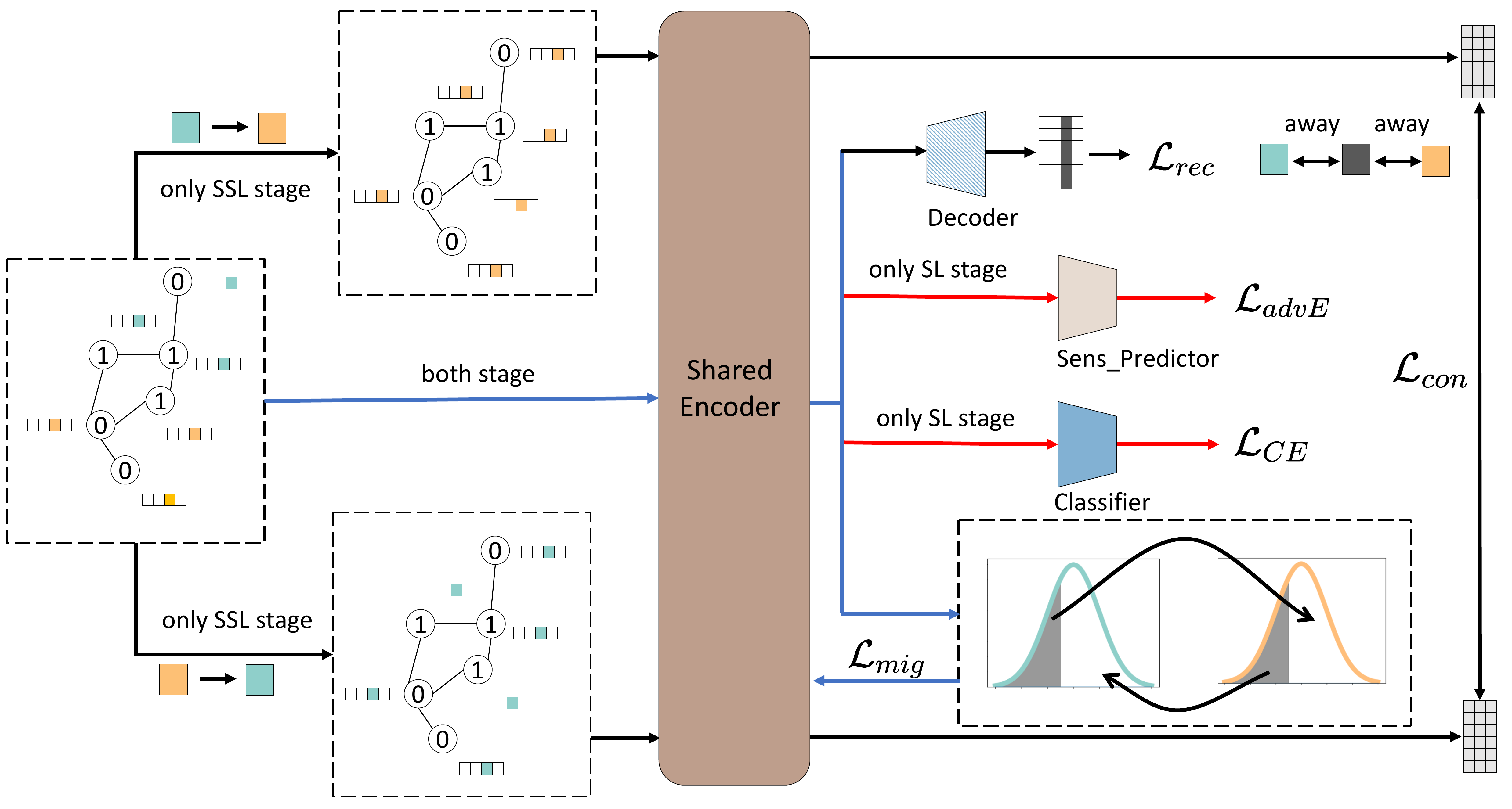}
    \caption{Framework of FairMigration. The workflow of the Self-Supervised Learning (SSL) stage only is colored black. 
    The workflow of the Supervised Learning (SL) stage only is colored red. 
    The workflow of both the SSL stage and SL stage is colored blue. 
    At the SSL stage, FairMigration optimizes the encoder by personalized self-supervised learning 
    and migrates the demographic groups.
    At the SL stage, FairMigration optimizes the encoder and the classifier under the constraints of migrated groups
    and adversarial training.
    }
    \label{fig:framework}
    
\end{figure*}

In this section, the proposed model, FairMigration will be introduced in detail. 
FairMigration consists of two training stages.
In the first stage (self-supervised learning stage),
FairMigration constructs pseudo-demographic groups by 
group migration based on similarity while initially optimizing 
the encoder using self-supervised learning based on counterfactual fairness.  
The division of pseudo-demographic groups is expected to correspond to the ground-truth labels, 
rather than the original sensitive attributes.
The procedure of regenerating new pseudo-demographic groups, conducted in a self-supervised manner, escapes from biased prediction.
The obtained pseudo-demographic groups are 
applied to the model's training in the supervised learning stage as a fairness constraint.
In the second stage (supervised learning stage), 
in addition to the cross-entropy, 
the adversarial training and 
the pseudo-demographic group based distance restriction 
are added to further promote the fairness issue. The pseudocode is demonstrated in Algorithm \ref{alg:label}, and the framework is displayed in Figure \ref{fig:framework}.

\begin{algorithm}
    \SetAlgoLined
    \KwIn{Original graph $\mathcal{G} (\mathbf{X}, \mathbf{A})$}
    \KwOut{Graph Encoder $GNN(\cdot ,\cdot)$, Label Predictor $MLP(\cdot)$}
    
    \caption{FairMigration}
    \label{alg:label}
    \tcp{SSL stage}\label{ssl} 
    \While{self-supervised learning}{
        Augmente the graph $\mathcal{G} (\mathbf{X}, \mathbf{A})$ based on 
        Counterfactual fairness;

        Calculate the \textbf{group similarity distribution};

        Sensitive attribute value flip for nodes meet with equation (\ref{deviation});

        Optimize $GNN(\cdot ,\cdot)$ by equation (\ref{pre_loss});
    }
    Freeze pseudo-demographic groups;
    
    \tcp{SL stage}\label{sl} 
    
    \While{supervised learning}{
        Optimize $GNN(\cdot ,\cdot)$ and $MLP(\cdot)$ by equation (\ref{sup_loss});
    }

    \Return $GNN(\cdot ,\cdot)$ and $MLP(\cdot)$;
\end{algorithm}

\subsection{Self-Supervised Learning Stage}
The target of FairMigration is to improve the fairness of the graph neural network 
through group migration. 
Augmentation of the sensitive attributes during the self-supervised learning stage 
meets with the downstream tasks. Also, The encoder can be optimized preliminarily.
Therefore, we choose sensitive attributes flip as the augmentation strategy for 
self-supervised training.

\subsubsection{Counterfactual Fairness Augmentation.}
Given a graph $\mathcal{G}(\mathbf{A},\mathbf{X})$, $\mathbf{A}$ is the adjacency matrix,
and $\mathbf{X}$ is the attribute matrx.
Two augmented views of $\mathcal{G}$, generated by 
setting all the sensitive attributes $\mathbf{S}$ as 0 and 1, can be annotated as
$\tilde{\mathcal{G}}_0(\mathbf{A},\tilde{\mathbf{X}}_0)$ and 
$\tilde{\mathcal{G}}_1(\mathbf{A},\tilde{\mathbf{X}}_1)$, respectively.
Such an augmentation strategy would encourage the graph neural networks 
to get rid of the false relationships between predicted labels 
$\tilde{\mathbf{Y}}$ and the sensitive attributes
$\mathbf{S}$.

Then, we use GNN to obtain the embeddings of 
$\mathcal{G}(\mathbf{A},\mathbf{X})$,
$\tilde{\mathcal{G}}_0(\mathbf{A},\tilde{\mathbf{X}}_0)$,
and $\tilde{\mathcal{G}}_1(\mathbf{A},\tilde{\mathbf{X}}_1)$.
The embedding matrix of graph ${\mathcal{G}}$ can be written as:
\begin{equation}
    \mathbf{Z} = GNN(\mathbf{A},\mathbf{X}).
    \label{ori_emb}
\end{equation}
The embedding matrix of augmented graph $\tilde{\mathcal{G}}_i$ can be written as:
\begin{equation}
    \tilde{\mathbf{Z}}_i = GNN(\mathbf{A},\tilde{\mathbf{X}}_i),\quad i = 0,1. 
    \label{aug_emb}
\end{equation}

And we defined a contrastive loss $\mathcal{L}_{con}$ to optimize the encoder for the fairness in equation (\ref{con_loss}),
where $shuffle(\cdot)$ is the shuffle function.
\begin{equation}
    \begin{aligned}
    \mathcal{L}_{con} =  \quad&\frac{1}{N} \sum_i^N  ((1-cos(\tilde{\mathbf{Z}}_1[i,:], 
    \tilde{\mathbf{Z}}_2[{i},:]))
    \\
    &
    + cos(\tilde{\mathbf{Z}}_1[i,:], shuffle(\tilde{\mathbf{Z}}_2[i,:]))).
    \end{aligned}
    \label{con_loss}
\end{equation}

Additionally, we introduce a personalized reconstruction loss $\mathcal{L}_{rec}$ to 
enhance the representation ability of the encoder
while blurring the sensitive attributes.
We adopt MLP as a decoder to reconstruct the attributes, 
avoiding sensitive attribute leakage caused by message passing:
\begin{equation}
    \mathbf{X}^{rec} = MLP(GNN(\mathbf{A},\mathbf{X})),
\end{equation}
\begin{equation}
    \mathcal{L}_{rec} = \frac{1}{N} \sum_i^N (  
    MSE(\mathbf{U},\mathbf{U}^{rec}) + \sum_{j=0}^{j=s_{max}} MSE(S^{rec}_i,j)),
    \label{rec_loss}
\end{equation}
where $MSE(\cdot,\cdot )$ is the mean-square error.
$\mathbf{U}$ is the attributes matrix without the sensitive attribute, 
$\mathbf{U}^{rec}$ is the reconstructed attributes matrix $\mathbf{X}^{rec}$ 
without the reconstructed sensitive attribute $\mathbf{S}^{rec}$, and $ cos(\cdot, \cdot)$ 
is cosine similarity.
Letting $q$ be the index of sensitive attribute channel in $\mathbf{X}$, $\mathbf{U}$ can be represented as  
$\mathbf{U} = [\mathbf{X}[:,0:q],\mathbf{X}[:,q+1:]]$.

The following optimization function is applied to optimize the GNN encoder in order to
stripping the prediction label from the original 
sensitive attribute $\mathbf{S}$.
\begin{equation}
    \mathcal{L}_{ssl} = \alpha \mathcal{L}_{con} + (1 - \alpha) \mathcal{L}_{rec},
    \label{ssl_loss}
\end{equation}
where $\alpha \in [0,1]$ is a hyperparameter that balances $\mathcal{L}_{con}$ and $\mathcal{L}_{rec}$.

\subsubsection{Demographic Groups Migration.}
While self-supervised learning, 
a similarity-based pseudo-demographic group migration is conducted
to constrain the GNN for fairness.
The group division can be adjusted dynamically when self-supervised learning.
Thus, group migration breaks the limitation of static sensitive attributes.

Given the current pseudo sensitive attribute vector $\mathbf{P}$, 
the prototype set of all pseudo-demographic groups 
$\{\mathbf{T}_0,\mathbf{T}_1, \cdot\cdot\cdot, 
\mathbf{T}_{s_{max}}\}$ can be acquired:
\begin{equation}
    \mathbf{T}_{j} = {\frac{1}{|\mathbf{P}_i = {j}|}}\sum_{i \in \{\mathbf{P}_i = {j}\}} \mathbf{Z}_i,
    \label{centerj}
\end{equation}
where $\mathbf{T}_j, j \in \{0, 1, \cdot\cdot\cdot, S_{max} \}$ (in this paper, $S_{max} = 1$) 
is the prototype of $j$-th group, and
$\mathbf{P}_i$ is the $i$-th node's current pseudo sensitive attribute.

After that, pseudo-demographic groups' similarities matrix $\mathbf{Q}$ can be calculated, 
which implies the similarity distribution of groups:
\begin{equation}
    \mathbf{Q}_i = cos(\mathbf{T}_{\mathbf{P}_i}, \mathbf{Z}_i), 
    \label{sim_score}
\end{equation}
where $\mathbf{Q}_i$ measures the cosine similarity of the $i$-th node with its 
current pseudo-demographic group's prototype.
The mean value $\mu_k$ and the standard deviation $ \sigma_k $ of set $\{\mathbf{Q}_i : \mathbf{P}_i = k \}$
describe the similarity distribution of the pseudo-demographic group $k$.

The outliers satisfying equation (\ref{deviation}) 
(deviating from group up to the threshold)
will be migrated to the group whose prototype is the most similar.
This paper is conducted on single-dimension binary sensitive attributes. 
Therefore, the group migration is simplified into a sensitive attribute value flip as
$\mathbf{P}_i = 1 - \mathbf{P}_i$.
The situation of generating new groups is not considered in this paper.
\begin{equation}
    \mathbf{Q}_i < \mu_{\mathbf{P}_i} - 2 \times \sigma_{\mathbf{P}_i}.
    \label{deviation}
\end{equation}

The loss function of group migration can be written as:
\begin{equation}
    \mathcal{L}_{mig} = \frac{1}{|\mathbf{O}|}\sum_{i \in \mathbf{O}} 1 - sim(\mathbf{Z}_i, 
    \mathbf{T}_{1-\mathbf{P}_i}).
    \label{mig}
\end{equation}

\subsubsection{Objective Function of Self-supervised Learning.}
It has been revealed that the fairness of model 
will be disturbed by balanced demographic group distribution \cite{FMP}.
To further promote the fairness of our model, 
a re-weight strategy correlated with the number of demographic groups is introduced 
to adjust the loss functions in
equation (\ref{con_loss}), equation (\ref{rec_loss}) and equation (\ref{mig}).
The above loss functions can be rewritten as:
\begin{equation}
    \begin{aligned}
        \mathcal{L}_{con} = \quad & \frac{1}{N} \sum_i^N w_i 
        ((1-cos(\tilde{\mathbf{z}}_1[i,:], \tilde{\mathbf{z}}_2[{i},:])) 
        \\
        &
        + cos(\tilde{\mathbf{z}}_1[i,:], shuffle (\tilde{\mathbf{z}}_2[i,:]))),
    \end{aligned}
    \label{con_loss_new}
\end{equation}

\begin{equation}
    \begin{aligned}
    \mathcal{L}_{rec} = \quad & \frac{1}{N} \sum_i^N  w_i(  
    MSE(\mathbf{U},\mathbf{U}^{rec})  
            \\
        &
    + \sum_{j=0}^{j=s_{max}} MSE(S^{rec}_i,j)
    ),
    \end{aligned}
    \label{rec_loss_new}
\end{equation}

\begin{equation}
    \mathcal{L}_{mig} = \frac{1}{|\mathbf{O}|}\sum_{i \in \mathbf{O}} w_i (1 - {sim(\mathbf{Z}_i, 
    \mathbf{T}_{1-\mathbf{P}_i})}),
    \label{mig_new}
\end{equation}
where $w_i = \frac{max\{|\mathbf{S} = 0|,|\mathbf{S} = 1|\} }{|\mathbf{S}_i|}$.

The loss function of self-supervised learning can be written as:
\begin{equation}
    \mathcal{L}_{pre} = \mathcal{L}_{mig} + \gamma \mathcal{L}_{ssl},
    \label{pre_loss}
\end{equation}
where $\gamma \in [0, 1]$ is hyperparameter to control the contribution of $\mathcal{L}_{ssl}$.

As self-supervised learning finishes, the migrated pseudo-sensitive attributes matrix $\mathbf{P}$
is frozen and will be used in the supervised learning phase as a fairness constraint.
\subsection{Supervised Learning Stage}

At the stage of self-supervised learning, only the encoder and decoder are 
optimized mainly for fairness.
Therefore, it is necessary to optimize the classifier and fine-tune the encoder 
for node classification.
In order to improve the fairness of the classifier and
avoid the encoder being undermined by the fairness reduction in supervised learning,
we introduce migrated pseudo-sensitive attributes and adversarial training as constraints.

\subsubsection{Cross-entropy Loss.}
A multilayer perceptron is adopted as a classifier to output the predicted labels 
$\tilde{\mathbf{Y}}$:
\begin{equation}
    \tilde{\mathbf{Y}} = \delta (MLP(GNN(\mathbf{A},\mathbf{X}))), 
    \label{Y_pre}
\end{equation}
where $\delta$ is the activation function.

The cross-entropy is used as the loss function of supervised learning:
\begin{equation}
    \mathcal{L}_{CE} = - \frac{1}{\mathbf{N}} \sum_i^{\mathbf{N}} 
    [y_i log(\tilde{y}_i) + (1-y_i)log(1 - \tilde{y}_i)].
    \label{loss_ce}
\end{equation}

\subsubsection{Pseudo-demographic Groups Constraints.}
In the supervised learning stage, 
the migration loss function in the self-supervised learning is still applied, 
but the migrated pseudo sensitive attributes are frozen. 
This procedure could prevent the fairness degradation brought by biased classification. 

\subsubsection{Adversarial Training.}
The adversarial training encourages the encoder to avoid exposing raw sensitive attributes while 
optimizing the sensitive attributes predictor.
In this paper, we set {all} sensitive attributes available. 
However, a portion of individuals may offer fake sensitive attributes for privacy.
We adopt a sensitive attributes predictor $f_E(\cdot)$ to recover the sensitive attributes:
\begin{equation}
    \mathbf{S}^{p} = f_E (\mathbf{A},\mathbf{X}),
    \label{sens_pre}
\end{equation}
where $\mathbf{S}^{p}$ is the recovered sensitive attributes matrix.

With $\mathbf{S}^{p}$, The optimization direction of adversarial training can be transferred 
to reducing sensitive information in embedding $\mathbf{Z}$.
The adversarial training module $f_A(\cdot)$ tries to retrieve the sensitive attributes from $\mathbf{Z}$
while the encoder tries to eliminate the sensitive attributes in $\mathbf{Z}$.
The retrieved sensitive attributes $\mathbf{S}^{A}$ from $f_A(\cdot)$ can be written as:
\begin{equation}
    \mathbf{S}^{A} = f_A (\mathbf{A},\mathbf{X}),
    \label{sens_ADV}
\end{equation}

The loss function of optimizing $f_E(\cdot)$ can be written as:
\begin{equation}
    \mathop{min}\limits_{\theta_E} \mathcal{L}_{E} = -\frac{1}{\mathbf{N}}
    [\sum_i \mathbf{S}^{A}_i log(\mathbf{S}^{p}_i) +
    (1 - \mathbf{S}^{A}_i) log(1 - \mathbf{S}^{p}_i)],
    \label{adv_loss}
\end{equation}
where $\theta_E$ is the trainable parameters of $f_E(\cdot)$.

To avoid the exposure of the raw sensitive attributes and the leaking of 
flipped sensitive attributes, 
instead of a simple cross-entropy function, the loss function of adversarial training 
is modified as:
\begin{equation}
    \begin{aligned}
    \mathop{min}\limits_{\theta_G} \mathop{max}\limits_{\theta_A} \mathcal{L}_{A} 
    =& -\frac{1}{2\mathbf{N}}
    [\sum_i \mathbf{S}^A_i log(\mathbf{S}^{p}_i) + 
    \\ 
    &
    (1 - \mathbf{S}^A_i) log(1 - \mathbf{S}^{p}_i)+ 
    \\ 
    &
    (1-\mathbf{S}^A_i) log(\mathbf{S}^{p}_i) + 
    \mathbf{S}^A_i log(1 - \mathbf{S}^{p}_i)],
    \end{aligned}
    \label{advE_loss}
\end{equation}   
where $\theta_A$ is the trainable parameters of $f_A(\cdot)$.

\subsubsection{Objective Function of Supervised Learning.}
The loss function of supervised learning can be written as:
\begin{equation}
    \left\{
        \begin{aligned}
            &\mathop{min}\limits_{\theta_G, \theta_C} \mathop{max}\limits_{\theta_A} \mathcal{L}_{sup} = 
    \mathcal{L}_{CE}  + 
    \lambda \mathcal{L}_{mig} - \beta \mathcal{L}_{A}, \\
    &\mathop{min}\limits_{\theta_E} \mathcal{L}_{E}.
        \end{aligned}
    \right.
    \label{sup_loss}
\end{equation}
where $\lambda $ and $\beta $ are hyperparameters.  
$\theta_G, \theta_C$ are trainable parameters of the encoder and classifier, respectively.

{
\subsection{Algorithm Complexity.}
Assuming the complexity of GNN is $\mathcal{O}(G)$, the number of nodes is $\mathbf{N}$, the number of demographic groups is 2 in this paper, the dimension of attribute is $K$, and the dimension of embedding is $d$. 

On the Self-Supervised Learning stage,
the computational complexity of graph augmentation is $\mathcal{O}(2N)$, the computational complexity of $\mathcal{L}_{ssl}$ is $\mathcal{O}((d+K)N)$.
the computational complexity of $\mathcal{L}_{mig}$ is $\mathcal{O}(dN)$. FairMigration is $\mathcal{O}((2+2d+K)N+G)$.

On the Supervised Learning stage, 
the computational complexities of $\mathcal{L}_{CE}$, $\mathcal{L}_{A}$ and $\mathcal{L}_{E}$ are all $\mathcal{O}(N)$. the computational complexity of $\mathcal{L}_{mig}$ is still $\mathcal{O}(dN)$.
FairMigration is $\mathcal{O}((3+d)N+G)$.

To Sum up, FairMigration is $\mathcal{O}((5+3d+k)N+2G)$.
}

\section{Experiment}
In this section, 
a series of experiments are conducted to demonstrate the effectiveness of our proposed model.

\begin{table*}[b]
    \centering
    \caption{Dataset Statistics.}
    
      \begin{tabular}{l|c|c|c}
      \toprule
      \textbf{Dataset} & \textbf{Credit} & \textbf{Bail}
      & \textbf{Income} \\
      \midrule
      \textbf{\#node} & 30,000 & 18,876 & 14,821 \\
      \textbf{\#edge} & 200,526 & 403,977 & 51,386 \\
      \textbf{\#feature} & 13    & 18    & 14 \\
      \textbf{sensitive attribute} & age   & white & race \\
      \textbf{predict attribute} & \makecell[c]{default/no default} & \makecell[c]{bail/no bail} & \makecell[c]{income $\textgreater$ 50/$\leq$50} \\
      \bottomrule
      \end{tabular}%
      
    \label{dataset_statistics}%
\end{table*}%

\begin{table*}[hbtp]
    \centering
    \caption{Comparison of node classification and fairness with the state-of-the-art
    baselines. The best and the runner-up are marked by bold and underline, respectively.
    $\uparrow$ denotes the higher the better, and  $\downarrow$ denotes the lower the better.}
    \scalebox{0.65}{
      \begin{tabular}{c|l|p{4.2em}ll|lll|lll|c}
        \toprule
        \multicolumn{2}{c}{\multirow{2}[2]{*}{}} & \multicolumn{3}{c|}{Credit} & \multicolumn{3}{c|}{Bail} & \multicolumn{3}{c|}{Income} & \multicolumn{1}{c}{\multirow{2}[2]{*}{avg rank}} \\
    \cmidrule{3-11}    
    
    \multicolumn{2}{c}{} & \multicolumn{1}{c}{AUC($\uparrow$)} & \multicolumn{1}{c}{$\Delta SP$($\downarrow$)} & \multicolumn{1}{c|}{$\Delta EO$($\downarrow$)} & \multicolumn{1}{c}{AUC($\uparrow$)} & \multicolumn{1}{c}{$\Delta SP$($\downarrow$)} & \multicolumn{1}{c|}{$\Delta EO$($\downarrow$)} & \multicolumn{1}{c}{AUC($\uparrow$)} & \multicolumn{1}{c}{$\Delta SP$($\downarrow$)} & \multicolumn{1}{c|}{$\Delta EO$($\downarrow$)} &  \\
    
        \midrule
              \multirow{7}[0]{*}{GCN} & vanilla & \multicolumn{1}{l}{\underline{69.87$\pm$0.03}} & 13.40$\pm$0.30 & 12.80$\pm$0.50 & 87.18$\pm$0.42 & 7.44$\pm$0.68 & 5.04$\pm$0.67 & \textbf{77.32$\pm$0.04} & 25.98$\pm$0.57 & 31.01$\pm$0.42 & 13.44  \\
              & NIFTY & \multicolumn{1}{l}{69.17$\pm$0.10} & 10.76$\pm$0.72 & 10.08$\pm$1.15 & 78.82$\pm$2.98 & \textbf{2.53$\pm$1.93} & \textbf{1.80$\pm$1.22} & 72.93$\pm$1.41 & 24.90$\pm$3.64 & 25.75$\pm$7.12 & 11.11  \\
              & FairGNN & \multicolumn{1}{l}{67.72$\pm$0.29} & 11.72$\pm$4.54 & 11.13$\pm$4.76 & \underline{87.69$\pm$0.63} & 6.49$\pm$0.44 & 4.02$\pm$0.51 & 75.22$\pm$0.27 & 18.68$\pm$5.72 & 22.15$\pm$5.84 & 11.67  \\
              & EDITS & \multicolumn{1}{l}{\textbf{70.41$\pm$0.51}} & \underline{10.73$\pm$2.16} & \underline{8.08$\pm$0.98} & 85.49$\pm$0.76 & 6.23$\pm$0.56 & 3.97$\pm$0.71 & 73.47$\pm$1.80 & 23.85$\pm$3.89 & 23.80$\pm$5.47 & \underline{10.22}  \\
              & SIND 1\% & \multicolumn{1}{l}
              {68.60$\pm$1.03} & 18.30$\pm$1.27 & 17.34$\pm$0.87 & 88.17$\pm$1.26 & 7.06$\pm$0.54 & 4.42$\pm$0.79 & 75.39$\pm$0.93 & 18.98$\pm$2.60 & 25.59$\pm$2.35 & 13.67  \\
              & SIND 10\% & \multicolumn{1}{l}
              {67.44$\pm$0.21} & 17.74$\pm$0.43 & 16.79$\pm$0.31 & 88.10$\pm$0.38 & 6.16$\pm$0.28 & 3.91$\pm$0.49 & 74.61$\pm$0.87 & \underline{16.53$\pm$1.07} & \underline{19.92$\pm$2.75} & 12.44  \\
              & FairMigration  & \multicolumn{1}{l}{69.24$\pm$0.26} & \textbf{6.13$\pm$2.09} & \textbf{4.87$\pm$2.25} & \textbf{88.26$\pm$0.31} & \underline{5.80$\pm$0.26} & \underline{3.63$\pm$0.45} & \underline{75.44$\pm$0.69} & \textbf{14.15$\pm$2.38} & \textbf{16.68$\pm$2.96} & \textbf{5.56}  \\
              
        \midrule
        
              \multirow{7}[0]{*}{JK} & vanilla & \multicolumn{1}{l}{\underline{70.12$\pm$0.51}} & 12.10$\pm$4.91 & 11.56$\pm$4.66 & {88.51$\pm$0.44} & 7.85$\pm$0.52 & 5.46$\pm$0.92 & \textbf{78.79$\pm$0.27} & 29.73$\pm$3.53 & 32.25$\pm$3.86 & 13.00  \\
              & NIFTY & \multicolumn{1}{l}{69.34$\pm$0.37} & 11.18$\pm$1.93 & 10.98$\pm$2.63 & 81.67$\pm$1.92 & \textbf{4.64$\pm$0.64} & \textbf{3.58$\pm$0.29} & 74.56$\pm$0.19 & {26.36$\pm$3.70} & {28.55$\pm$2.44} & 11.78  \\
              & FairGNN & \multicolumn{1}{l}{66.12$\pm$1.15} & 11.56$\pm$5.17 & {10.58$\pm$5.27} & \textbf{90.17$\pm$0.62} & 7.58$\pm$0.67 & \underline{3.85$\pm$1.13} & \underline{77.03$\pm$0.44} & 28.05$\pm$5.13 & 32.41$\pm$4.89 & 12.56  \\
              & EDITS & \multicolumn{1}{l}{\textbf{71.40$\pm$1.13}} & 18.15$\pm$11.13 & 16.50$\pm$12.08 & 86.46$\pm$1.25 & 10.67$\pm$2.18 & 8.54$\pm$3.09 & 75.43$\pm$1.58 & 26.85$\pm$6.86 & 28.69$\pm$8.91 & 15.89  \\
              & SIND 1\% & \multicolumn{1}{l}{65.54$\pm$0.85} & \underline{10.51$\pm$0.43} & \underline{9.83$\pm$1.48} & \underline{89.32$\pm$1.30} & 7.99$\pm$0.65 & 6.62$\pm$0.51 & 74.54$\pm$0.56 & \underline{14.85$\pm$1.46} & \underline{17.37$\pm$1.55} & \underline{10.78}  \\
              & SIND 10\% & \multicolumn{1}{l}{66.67$\pm$3.12} & 14.81$\pm$1.38 & 12.76$\pm$0.92 & 89.27$\pm$0.79 & 8.49$\pm$0.79 & 7.31$\pm$1.16 & 74.09$\pm$0.84 & \textbf{12.86$\pm$2.84} & \textbf{16.65$\pm$2.19} & 12.44  \\            
              & FairMigration  & \multicolumn{1}{l}{68.91$\pm$0.43} & \textbf{8.76$\pm$1.72} & \textbf{7.59$\pm$1.81} & 88.25$\pm$1.24 & \underline{7.09$\pm$0.88} & 3.93$\pm$0.85 & 76.85$\pm$1.42 & {25.32$\pm$3.14} & {26.73$\pm$5.23} & \textbf{9.89}  \\
              
        \midrule
        
              \multirow{7}[0]{*}{APPNP} & vanilla & \multicolumn{1}{l}{\underline{71.34$\pm$0.10}} & 13.51$\pm$1.16 & 13.07$\pm$1.20 & 87.72$\pm$0.30 & 5.25$\pm$0.74 & 3.78$\pm$0.62 & \textbf{84.09$\pm$0.16} & 24.85$\pm$0.65 & 25.25$\pm$2.00 & 9.22  \\
              & NIFTY & \multicolumn{1}{l}{70.38$\pm$0.38} & 10.54$\pm$1.21 & {9.29$\pm$1.28} & 81.45$\pm$0.85 & \underline{4.29$\pm$0.87} & \underline{3.20$\pm$0.83} & 75.39$\pm$0.55 & 30.37$\pm$1.67 & 31.99$\pm$3.36 & 10.78  \\
              & FairGNN & \multicolumn{1}{l}{69.73$\pm$0.89} & 15.28$\pm$6.78 & 14.17$\pm$6.55 & \textbf{89.10$\pm$0.51} & 5.88$\pm$0.73 & 5.18$\pm$0.48 & \underline{80.23$\pm$0.92} & \textbf{15.27$\pm$5.47} & \textbf{15.76$\pm$6.54} & 8.67  \\
              & EDITS & \multicolumn{1}{l}{\textbf{72.25$\pm$0.64}} & 11.53$\pm$2.42 & 9.34$\pm$1.61 & 87.78$\pm$1.14 & 15.50$\pm$6.84 & 12.88$\pm$3.80 & 79.71$\pm$0.25 & 23.17$\pm$0.32 & 31.69$\pm$1.21 & 11.56  \\
            & SIND 1\% & \multicolumn{1}{l}{68.53$\pm$0.69} & {16.33$\pm$3.49} & {15.97$\pm$2.62} & 89.92 $\pm$ 0.36& 9.35$\pm$0.78 & 8.22$\pm$1.14 & 78.05$\pm$1.16 & {24.57$\pm$6.65} & {30.75$\pm$5.99} &14.11  \\
              & SIND 10\% & \multicolumn{1}{l}{69.05$\pm$0.94} & \underline{8.94$\pm$0.39} & \underline{7.73$\pm$0.59} & 89.71$\pm$1.26 & 7.87$\pm$0.61 & 7.95$\pm$0.82 & 78.19$\pm$2.15& {22.88$\pm$5.16} & 28.35$\pm$4.48 & \underline{7.89}  \\
              & FairMigration  & 70.32$\pm$0.58 & \textbf{8.02$\pm$1.75} & \textbf{7.02$\pm$1.48} & \underline{88.85$\pm$0.34} & \textbf{0.95$\pm$0.63} & \textbf{1.70$\pm$0.71} & 79.04$\pm$2.83 & \underline{21.99$\pm$7.19} & \underline{24.43$\pm$6.93} & \textbf{4.33}  \\

        \bottomrule
        \end{tabular}}
    \label{tab:main_comparision}%
  \end{table*}%

\begin{table*}[h]
\centering
\caption{{The investigation of message propagation on fairness.}}
\scalebox{0.7}{
    {\begin{tabular}{c|l|p{4.1em}ll|lll|lll}
    \toprule
    \multicolumn{2}{c}{\multirow{2}[4]{*}{}} & \multicolumn{3}{c|}{Credit} & \multicolumn{3}{c|}{Bail} & \multicolumn{3}{c}{Income} \\
\cmidrule{3-11}    \multicolumn{2}{c}{} & \multicolumn{1}{c}{AUC($\uparrow$)} & \multicolumn{1}{c}{$\Delta SP$($\downarrow$)} & $\Delta EO$($\downarrow$) & \multicolumn{1}{c}{AUC($\uparrow$)} & \multicolumn{1}{c}{$\Delta SP$($\downarrow$)} & $\Delta EO$($\downarrow$) & \multicolumn{1}{c}{AUC($\uparrow$)} & \multicolumn{1}{c}{$\Delta SP$($\downarrow$)} & $\Delta EO$($\downarrow$) \\
    \midrule
    \multirow{2}[2]{*}{GCN} & vanilla & 69.87$\pm$0.03 & 13.40$\pm$0.30 & 12.80$\pm$0.50 & 87.18$\pm$0.42 & 7.44$\pm$0.68 & 5.04$\pm$0.67 & 77.32$\pm$0.04 & 25.98$\pm$0.57 & 31.01$\pm$0.42 \\
            & FairMigration  & \multicolumn{1}{l}{{69.24$\pm$0.26}} & {6.13$\pm$2.09} & {4.87$\pm$2.25} & 88.26$\pm$0.31 & {5.80$\pm$0.26} & 3.63$\pm$0.45 & 75.44$\pm$0.69 & {14.15$\pm$2.38} & {16.68$\pm$2.96} \\
    \midrule
    \multirow{2}[2]{*}{JK} & vanilla & {70.12$\pm$0.51} & 12.10$\pm$4.91 & 11.56$\pm$4.66 & {88.51$\pm$0.44} & 7.85$\pm$0.52 & 5.46$\pm$0.92 & {78.79$\pm$0.27} & 29.73$\pm$3.53 & 32.25$\pm$3.86 \\
            & FairMigration  & \multicolumn{1}{l}{{68.91$\pm$0.43}} & {8.76$\pm$1.72} & {7.59$\pm$1.81} & {88.25$\pm$1.24} & {7.09$\pm$0.88} & {3.93$\pm$0.85} & 76.85$\pm$1.42 & {25.32$\pm$3.14} & {26.73$\pm$5.23} \\
    \midrule
    \multirow{2}[2]{*}{APPNP} & vanilla & {71.34$\pm$0.10} & 13.51$\pm$1.16 & 13.07$\pm$1.20 & 87.72$\pm$0.30 & 5.25$\pm$0.74 & 3.78$\pm$0.62 & {84.09$\pm$0.16} & 24.85$\pm$0.65 & 25.25$\pm$2.00 \\
            & FairMigration  & {70.32$\pm$0.58} & {8.02$\pm$1.75} & {7.02$\pm$1.48} & {88.85$\pm$0.34} & {0.95$\pm$0.63} & 1.70$\pm$0.71 & 79.04$\pm$2.83 & {21.99$\pm$7.19} & {24.43$\pm$6.93} \\

                          \midrule
                  \multicolumn{2}{c|}{{MLP}} & \multicolumn{1}{c}{69.93±0.20} & \multicolumn{1}{c}{9.91 ± 0.61} & \multicolumn{1}{c|}{9.25±0.84} & \multicolumn{1}{c}{87.28±0.49} & \multicolumn{1}{c}{2.65±0.44} & \multicolumn{1}{c|}{3.95±1.05} & \multicolumn{1}{c}{76.56±1.09} & \multicolumn{1}{c}{24.73±0.66} & \multicolumn{1}{c}{20.16±0.73}  \\
    \bottomrule
    \end{tabular}}}%
\label{tab:message}%
\end{table*}%

\subsection{Datasets}
We conduct experiments on three different real-world datasets
credit, bail, and income. 
The statistics information of the the datasets is demonstrated in Table \ref{dataset_statistics}.
The detailed introductions of these three datasets are as follows:
\begin{itemize}
    \item \textbf{Credit} \cite{credit}: credit graph is built on 30, 000 credit card users. 
    A node represents a user and the probability of generating an edge 
    between two nodes depends on the similarity of their payment features. 
    The label of credit is whether or not 
    the user will default on credit card payments next month.
    The sensitive attribute is age.
    \item \textbf{Bail} \cite{bail}: bail (also known as recidivism) graph is built on 18, 876 
    defendants released
    on bail at the U.S. state courts from 1990 to 2009.
    A node represents a defendant and the probability of generating an edge 
    between two nodes is determined by the similarity of their 
    past criminal records and demographics. 
    The label of bail is whether or not the defendant with bail.
    The sensitive attribute is race (white or not).
    \item \textbf{Income}: income graph is built on 14,821 individuals sampled from the 
    Adult dataset \cite{adult}.
    A node represents a person. Take the similarity of a pair of nodes as the probability 
    of establishing an edge between them.
    The label of income is whether or not the person earns more than 50K dollars.
    The sensitive attribute is race.
\end{itemize}

\subsection{Baselines}
In order to verify the effectiveness of FairMigration, 
three state-of-the-art GNN-based methods {proposed for node classification} NIFTY, FairGNN, EDITS, and SIND 
are chosen for comparison.
A brief introduction of these methods is following:
\begin{itemize}
    \item \textbf{NIFTY} \cite{nifty}. NIFTY promotes fairness by the 
    counterfactual perturbation-based siamese network and uses
    Lipschitz continuous function to normalize the layer weights. 
    \item \textbf{FairGNN} \cite{FAIRGNN}. FairGNN is an adversarial training-based method. 
    It trains a sensitive attributes predictor to retrieve 
    the sensitive attributes from the node embeddings
    while training the GNN to reduce the information 
    of the sensitive attributes in the node embeddings.
    We set all the sensitive attributes available for comparison.
    \item \textbf{EDITS} \cite{Edits}. EDITS transforms the attributes and 
    regenerates the adjacency matrix for
    lowering the Wasserstein distance of attributes 
    and topology between different demographic groups.
    \item \textbf{SIND} \cite{SIND}. SIND proposes a new metric to estimate the influence of nodes on fairness and alleviates the fairness issue by reducing the Wasserstein-1 distance of probability distribution between different demographic groups. The variants SIND 1\% and SIND 10\% ($k$\% denotes the budget of deleting nodes) provided by the authors are taken as baselines. 
\end{itemize}

\begin{table*}[b]
\centering
\caption{Ablation study. Comparison of FairMigration and its variants.}
\scalebox{0.7}{
    \begin{tabular}{c|l|p{4.1em}ll|lll|lll}
    \toprule
    \multicolumn{2}{c}{\multirow{2}[4]{*}{}} & \multicolumn{3}{c|}{Credit} & \multicolumn{3}{c|}{Bail} & \multicolumn{3}{c}{Income} \\
\cmidrule{3-11}    \multicolumn{2}{c}{} & \multicolumn{1}{c}{AUC($\uparrow$)} & \multicolumn{1}{c}{$\Delta SP$($\downarrow$)} & $\Delta EO$($\downarrow$) & \multicolumn{1}{c}{AUC($\uparrow$)} & \multicolumn{1}{c}{$\Delta SP$($\downarrow$)} & $\Delta EO$($\downarrow$) & \multicolumn{1}{c}{AUC($\uparrow$)} & \multicolumn{1}{c}{$\Delta SP$($\downarrow$)} & $\Delta EO$($\downarrow$) \\
    \midrule
    \multirow{5}[2]{*}{GCN} & w/o mig & \textbf{69.30$\pm$0.15} & 10.79$\pm$0.94 & 10.24$\pm$1.05 & \textbf{88.88$\pm$0.10} & 6.30$\pm$0.07 & 3.65$\pm$0.11 & 75.72$\pm$0.53 & 27.06$\pm$1.29 & 31.79$\pm$1.28 \\
            & w/o adv & \multicolumn{1}{l}{69.07$\pm$0.45} & 9.35$\pm$2.30 & 8.50$\pm$2.43 & 88.34$\pm$0.28 & 6.26$\pm$0.30 & 3.91$\pm$0.36 & \textbf{76.30$\pm$0.52} & 19.00$\pm$2.18 & 23.20$\pm$2.83 \\
            & w/o ssf & 69.00$\pm$0.36 & 10.69$\pm$2.06 & 10.12$\pm$2.40 & 87.93$\pm$0.21 & 6.13$\pm$0.10 & \textbf{3.31$\pm$0.29} & \underline{76.02$\pm$0.92} & 21.11$\pm$3.42 & 25.38$\pm$4.09 \\
            & w/o wei & \multicolumn{1}{l}{69.23$\pm$0.22} & \underline{6.70$\pm$2.56} & \underline{5.71$\pm$2.89} & \underline{88.41$\pm$0.34} & \underline{5.93$\pm$0.21} & \underline{3.53$\pm$0.36} & 75.21$\pm$0.42 & \textbf{13.56$\pm$2.43} & \underline{16.69$\pm$3.34} \\
            & FairMigration  & \multicolumn{1}{l}{\underline{69.24$\pm$0.26}} & \textbf{6.13$\pm$2.09} & \textbf{4.87$\pm$2.25} & 88.26$\pm$0.31 & \textbf{5.80$\pm$0.26} & 3.63$\pm$0.45 & 75.44$\pm$0.69 & \underline{14.15$\pm$2.38} & \textbf{16.68$\pm$2.96} \\
    \midrule
    \multirow{5}[2]{*}{JK} & w/o mig & \multicolumn{1}{l}{68.75$\pm$0.54} & 11.00$\pm$1.45 & 10.11$\pm$1.58 & \textbf{88.36$\pm$0.63} & 8.67$\pm$0.81 & 4.20$\pm$0.83 & \textbf{79.02$\pm$0.23} & 28.99$\pm$1.03 & 30.88$\pm$1.05 \\
            & w/o adv & \multicolumn{1}{l}{\textbf{68.99$\pm$0.77}} & 11.15$\pm$2.10 & 10.32$\pm$2.12 & 88.03$\pm$0.88 & 7.73$\pm$0.85 & \underline{3.96$\pm$0.66} & 76.89$\pm$1.62 & \textbf{24.54$\pm$3.91} & \underline{26.89$\pm$5.83} \\
            & w/o ssf & \multicolumn{1}{l}{68.56$\pm$0.42} & 11.01$\pm$2.45 & 10.25$\pm$2.52 & 87.20$\pm$1.01 & {7.59$\pm$0.61} & {4.81$\pm$1.06} & \underline{78.56$\pm$0.63} & 26.61$\pm$1.37 & 30.04$\pm$1.97 \\
            & w/o wei & \multicolumn{1}{l}{68.73$\pm$0.72} & \underline{8.85$\pm$3.77} & \underline{7.78$\pm$3.89} & 88.13$\pm$1.30 & \underline{7.52$\pm$0.92} & 4.18$\pm$0.67 & 77.53$\pm$0.59 & 26.24$\pm$1.96 & 29.56$\pm$3.12 \\
            & FairMigration  & \multicolumn{1}{l}{\underline{68.91$\pm$0.43}} & \textbf{8.76$\pm$1.72} & \textbf{7.59$\pm$1.81} & \underline{88.25$\pm$1.24} & \textbf{7.09$\pm$0.88} & \textbf{3.93$\pm$0.85} & 76.85$\pm$1.42 & \underline{25.32$\pm$3.14} & \textbf{26.73$\pm$5.23} \\
    \midrule
    \multirow{5}[2]{*}{APPNP} & w/o mig & \multicolumn{1}{l}{\textbf{70.38$\pm$0.39}} & 11.61$\pm$2.06 & 10.66$\pm$2.05 & 88.59$\pm$0.31 & 1.82$\pm$0.60 & 2.24$\pm$0.39 & \underline{81.18$\pm$0.32} & 28.63$\pm$0.55 & 32.45$\pm$0.80 \\
            & w/o adv & \multicolumn{1}{l}{68.15$\pm$2.19} & 8.44$\pm$4.43 & 7.98$\pm$4.45 & 88.70$\pm$0.23 & \textbf{0.67$\pm$0.54} & \textbf{1.34$\pm$0.52} & 79.36$\pm$1.82 & \textbf{21.24$\pm$3.53} & \textbf{22.02$\pm$4.36} \\
            & w/o ssf & 70.03$\pm$0.26 & 10.03$\pm$1.04 & 9.32$\pm$1.22 & \textbf{88.86$\pm$0.49} & 1.06$\pm$0.69 & \underline{1.53$\pm$0.72} & \textbf{81.24$\pm$0.43} & 23.81$\pm$2.25 & 26.72$\pm$2.99 \\
            & w/o wei & \multicolumn{1}{l}{66.08$\pm$4.46} & \textbf{6.28$\pm$4.64} & \textbf{6.04$\pm$3.76} & 88.65$\pm$0.39 & 1.12$\pm$0.85 & 1.81$\pm$0.58 & 79.58$\pm$1.81 & 23.99$\pm$4.35 & 25.85$\pm$5.32 \\
            & FairMigration  & \underline{70.32$\pm$0.58} & \underline{8.02$\pm$1.75} & \underline{7.02$\pm$1.48} & \underline{88.85$\pm$0.34} & \underline{0.95$\pm$0.63} & 1.70$\pm$0.71 & 79.04$\pm$2.83 & \underline{21.99$\pm$7.19} & \underline{24.43$\pm$6.93} \\

    \bottomrule
    \end{tabular}}%
\label{tab:ABLATION}%
\end{table*}%

\begin{figure*}
    \centering
    \subfigure[AUC]{
        \includegraphics[scale=0.4]{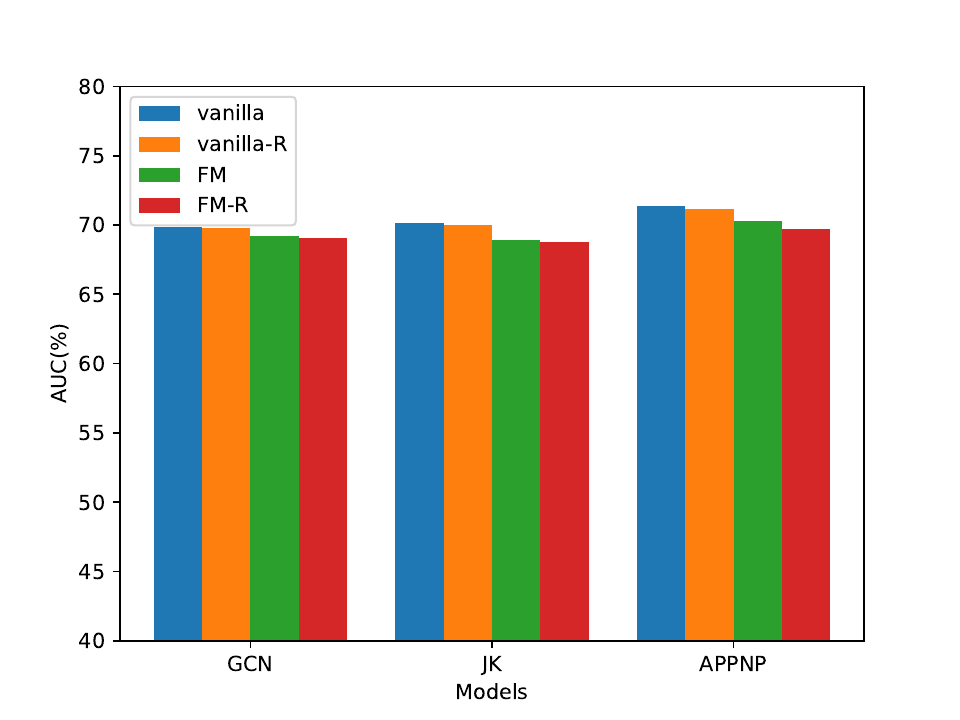}
        \label{AUC_sens_flip}
    }
    \subfigure[$\Delta$SP]{
        \includegraphics[scale=0.4]{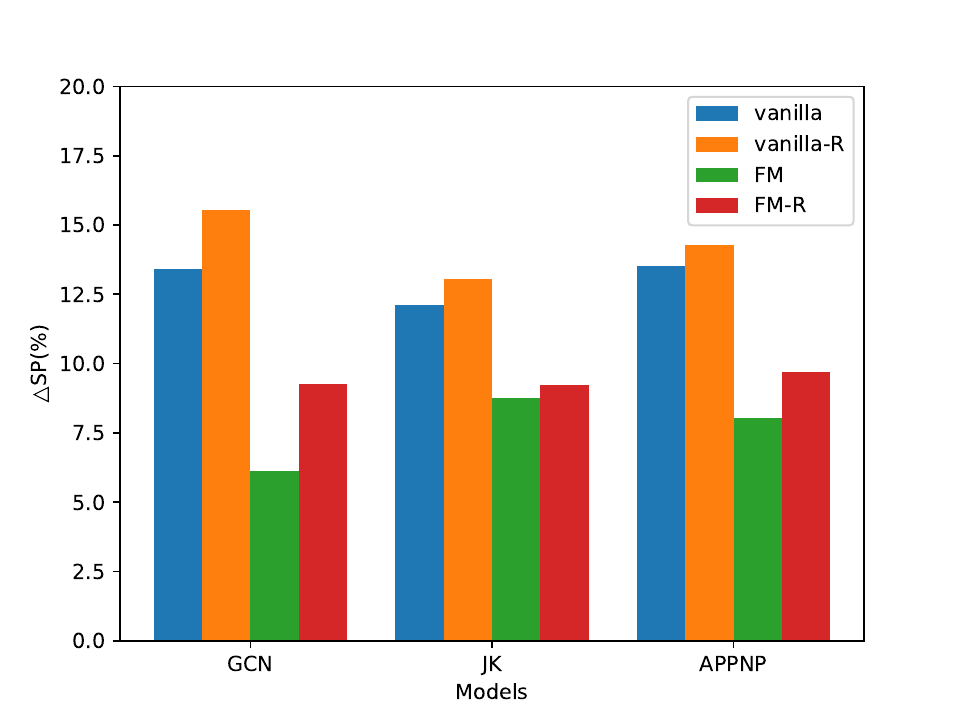}
        \label{SP_sens_flip}
    }
    \caption{{Flipping the sensitive attribute (age) on credit when testing. The vanilla denotes testing the vanilla GNNs without sensitive attribute flipping. The vanilla-R denotes testing the vanilla GNNs with sensitive attribute flipping. The FM denotes testing FairMigration without sensitive attribute flipping. The FM-R denotes testing FairMigration with sensitive attribute flipping.}  }
    \label{sens_flip}
\end{figure*}

\begin{figure*}[htbp]
    \centering
    \subfigure[Mean on credit]{
        \includegraphics[scale=0.3]{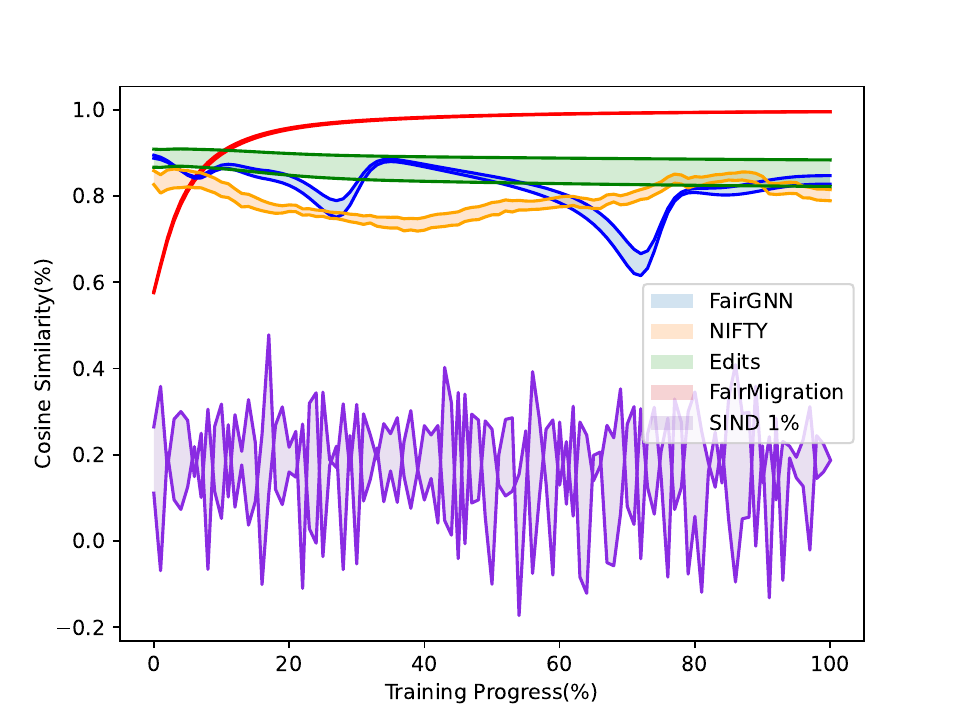}
        \label{ssf_gcn_credit_mean}
    }
    \subfigure[Mean on bail]{
        \includegraphics[scale=0.3]{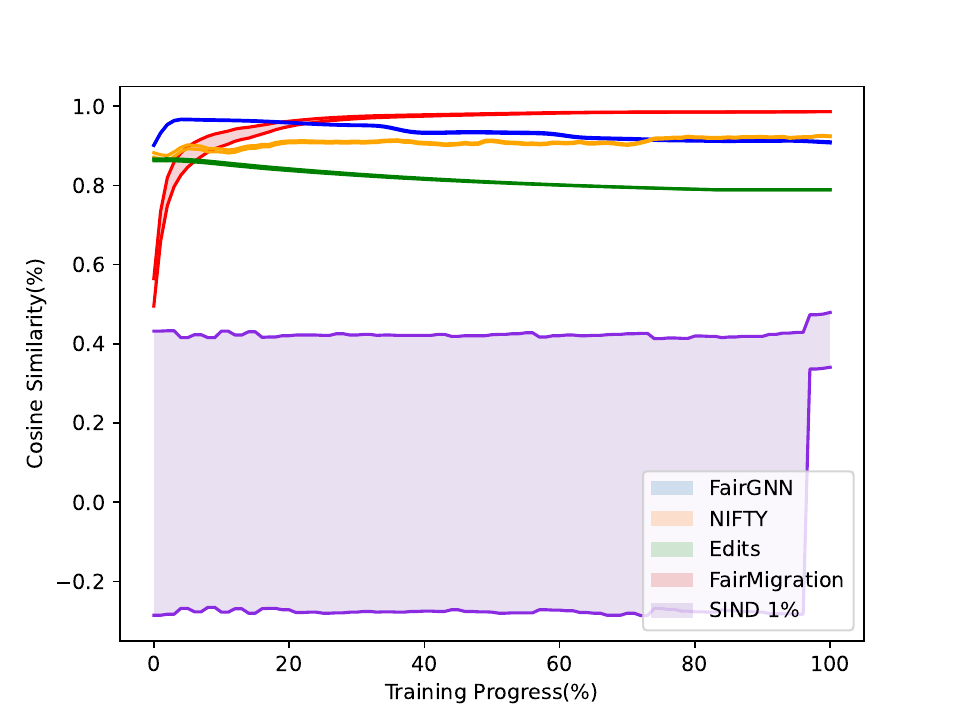}
        \label{ssf_gcn_credit_mean}
    }
    \subfigure[Mean income]{
        \includegraphics[scale=0.3]{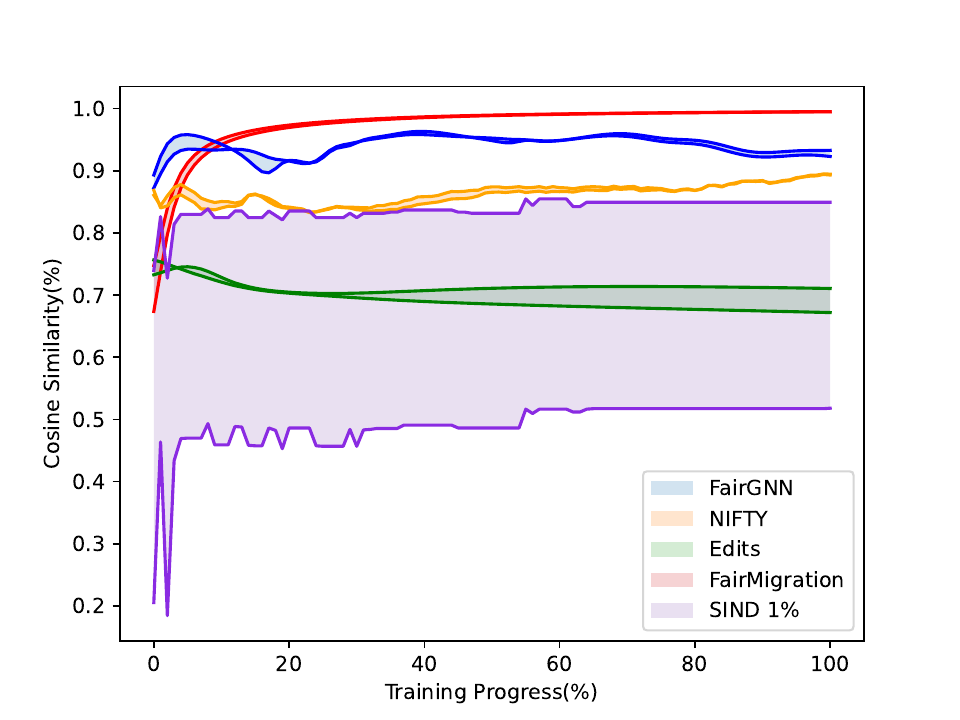}
        \label{ssf_gcn_credit_mean}
    }

    \subfigure[Std on credit]{
        \includegraphics[scale=0.3]{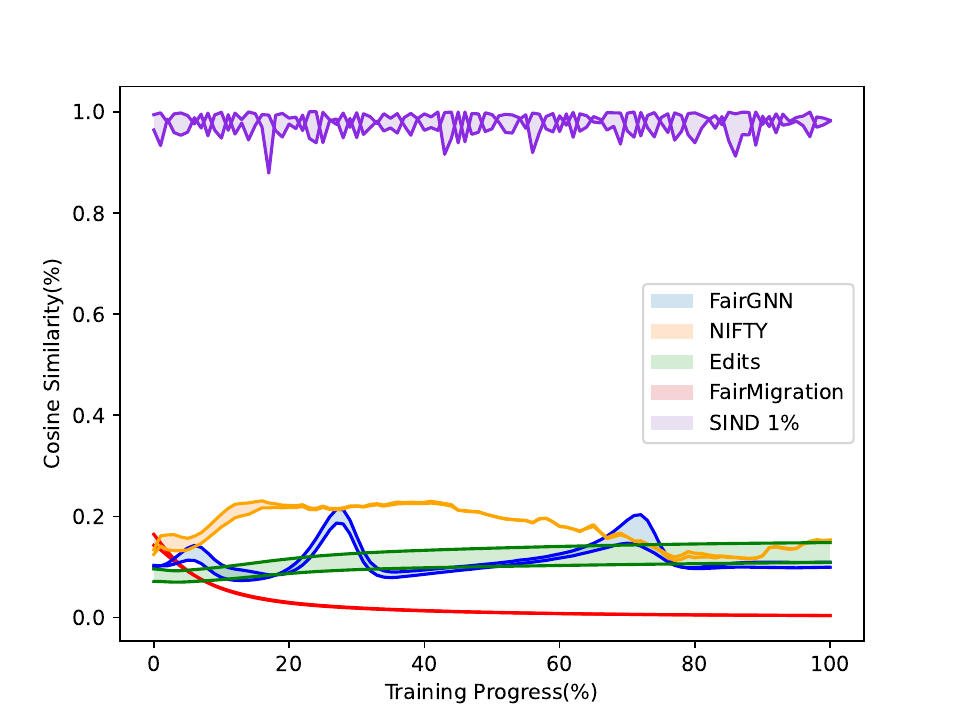}
        \label{ssf_gcn_credit_std}
    }
    \subfigure[Std on bail]{
        \includegraphics[scale=0.3]{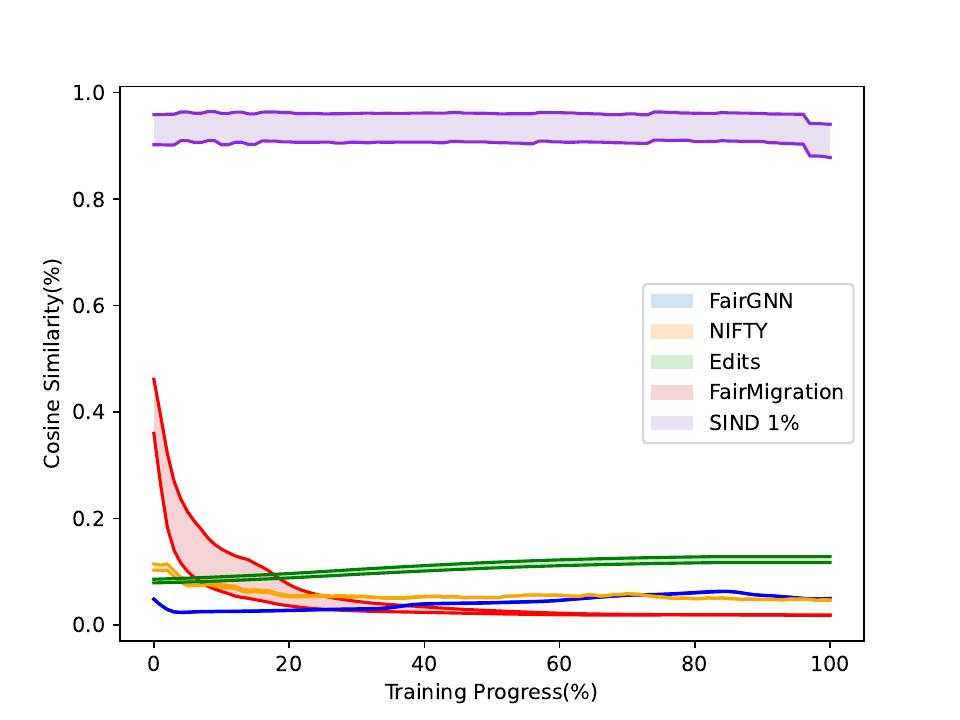}
        \label{ssf_gcn_credit_std}
    }
    \subfigure[Std on income]{
        \includegraphics[scale=0.3]{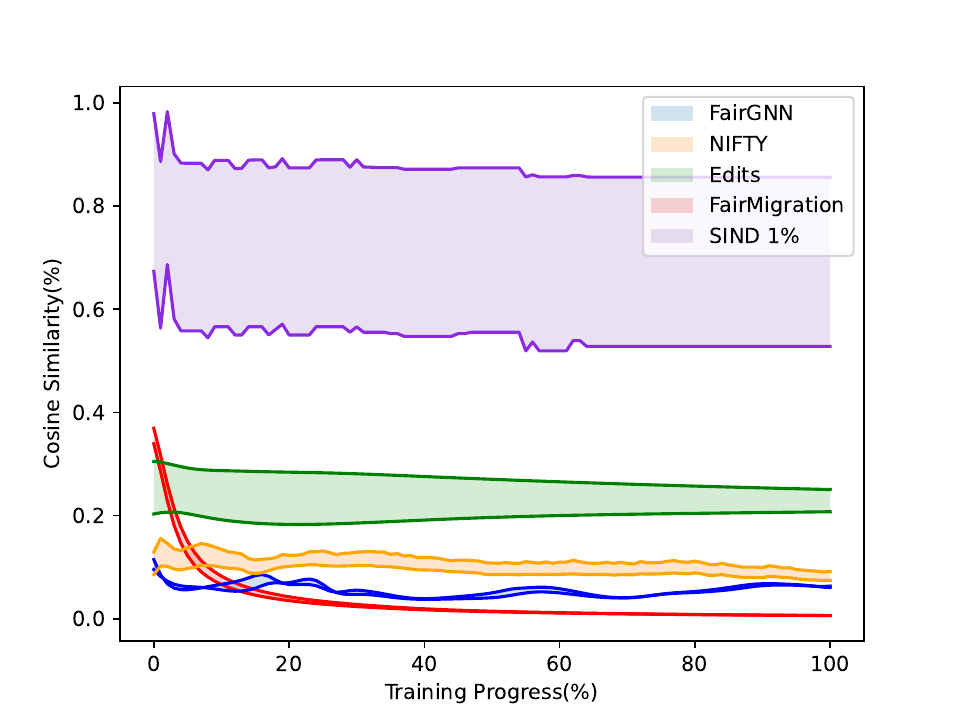}
        \label{ssf_gcn_credit_std}
    }
    \caption{Visualization of group migration. The bold lines with the
    same color indicate the mean/std of two group similarity distributions from the same method, respectively. 
    The shaded areas indicate the group similarity distribution difference. 
    The wider the shaded area on the y-axis, the larger similarity difference between groups.}
    \label{vis_mig}
\end{figure*}

\subsection{Evaluation Metrics}
We use AUC-ROC to evaluate the performance of node classification. 
In addition, we use statistical parity($\Delta SP$, also known as demographic parity) and 
equal opportunity($\Delta EO$) to evaluate the fairness.
The definition of $\Delta SP$ and $\Delta EO$ can be written as:
\begin{equation}
    \Delta SP = |P(\hat{y} = 1|s = 0) - P(\hat{y} = 1|s = 1)|,
    \label{sp}
\end{equation}
\begin{equation}
\begin{aligned}
    \Delta EO = \quad\quad |P(\hat{y} = 1|y=1,s = 0) -
    &\\ 
    P(\hat{y} = 1|y=1,s = 1)|.
\end{aligned}
    \label{eo}
\end{equation}

\subsection{Implementation}
The experiments are conducted on a server with Intel(R) Core(TM) 
i9-10980XE CPU @ 3.00GHz, NVIDIA 3090ti, 
Ubuntu 20.04 LTS,
CUDA 11.3, python3.8, PyTorch 1.12.1, and PyTorch Geometric.
Three popular GNNs, GCN, JK, and APPNP are adopted to be the backbone.
The parameters range of experiments are following:
\begin{itemize}
    \item \textbf{FairGNN}: $\alpha=4, \beta \in \{10, 100, 1000\}$.
    \item \textbf{NIFTY}: $\lambda = 0.5$.
    \item \textbf{Edits}: Default $u_1$ to $u_4$. Threshold = 0.02 for credit, 
    0.015 for bail, 0.1 for income.
    \item \textbf{SIND}: Default setting.
    \item \textbf{FairMigration}: $\lambda \in\{5,10,15,20\}$,\\  
$\beta \in \{0.01, 0.05,0.1, 0.5, 1\}$, 
$\alpha \in \{0.2, 0.4, 0.6,0.8\}$, 
$\gamma \in \{0.2, 0.4, 0.6, 0.8, 1\}$.
\end{itemize}
We run all experiments 10 times to prevent accidents as much as possible.

\begin{figure*}[t]
    \centering
    \subfigure[Impact of $\lambda$ to AUC]{
        \includegraphics[scale=0.25]{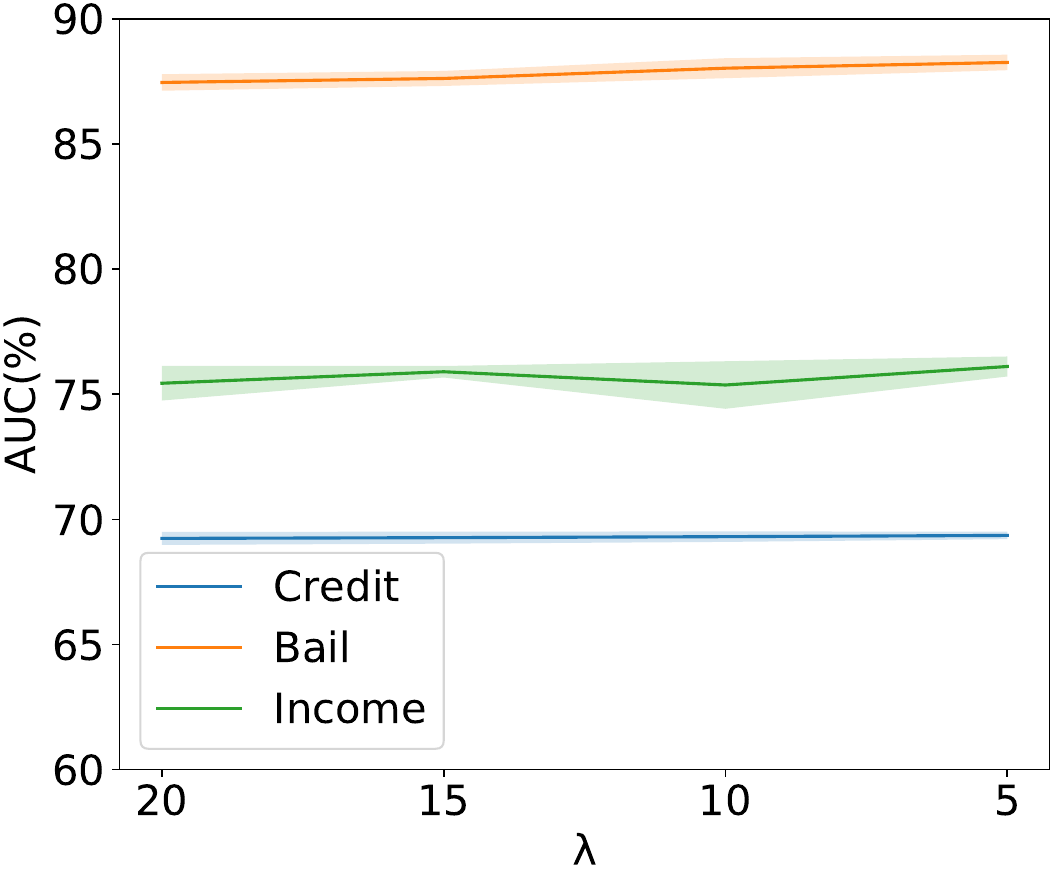}
        \label{lamb_auc}
    }
    \subfigure[Impact of $\lambda$ to $\Delta SP$]{
        \includegraphics[scale=0.25]{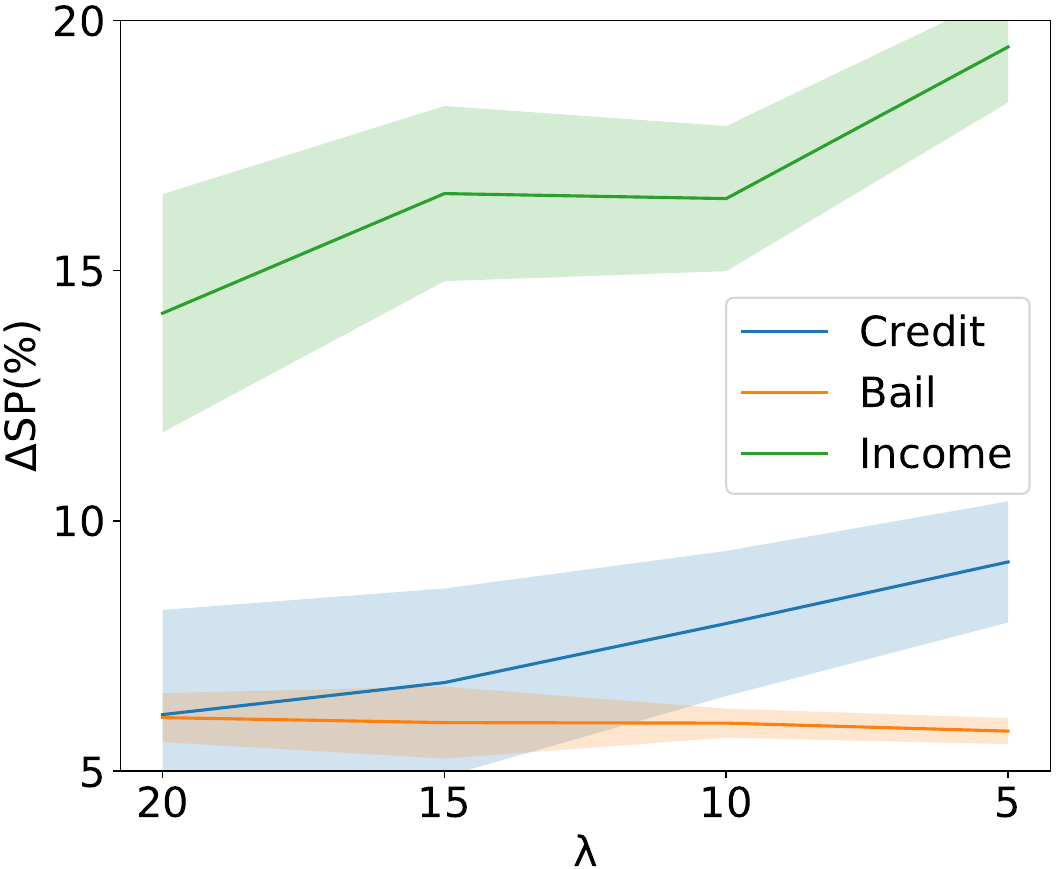}
        \label{lamb_sp}
    }
    \subfigure[Impact of $\lambda$ to $\Delta EO$]{
        \includegraphics[scale=0.25]{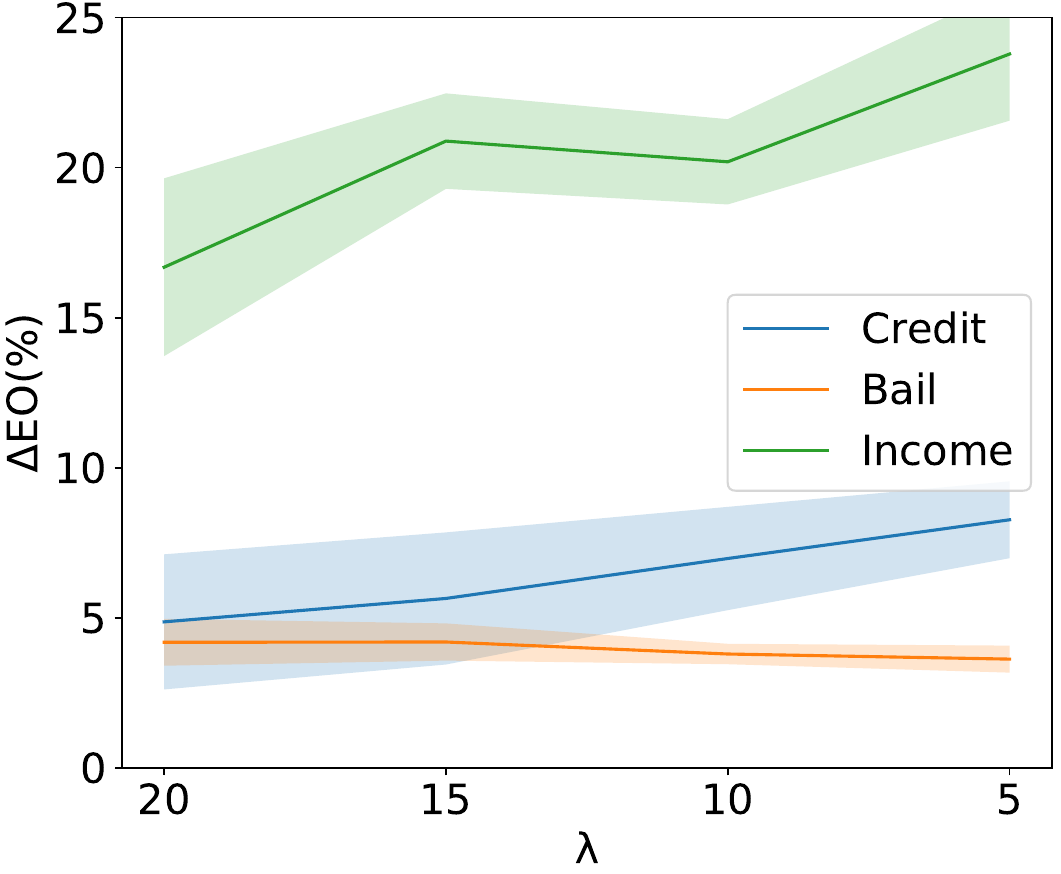}
        \label{lamb_eo}
    }

    \subfigure[Impact of $\beta$ to AUC]{
        \includegraphics[scale=0.25]{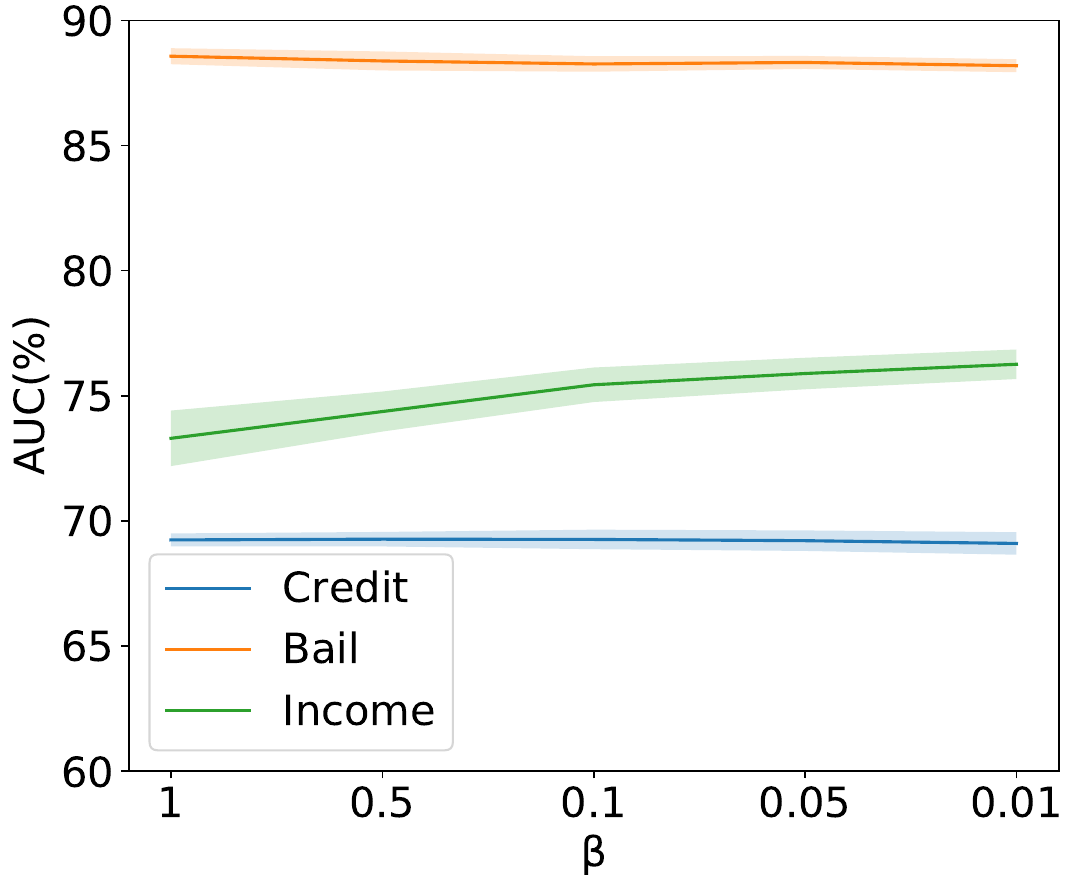}
        \label{beta_auc}
    }
    \subfigure[Impact of $\beta$ to $\Delta SP$]{
        \includegraphics[scale=0.25]{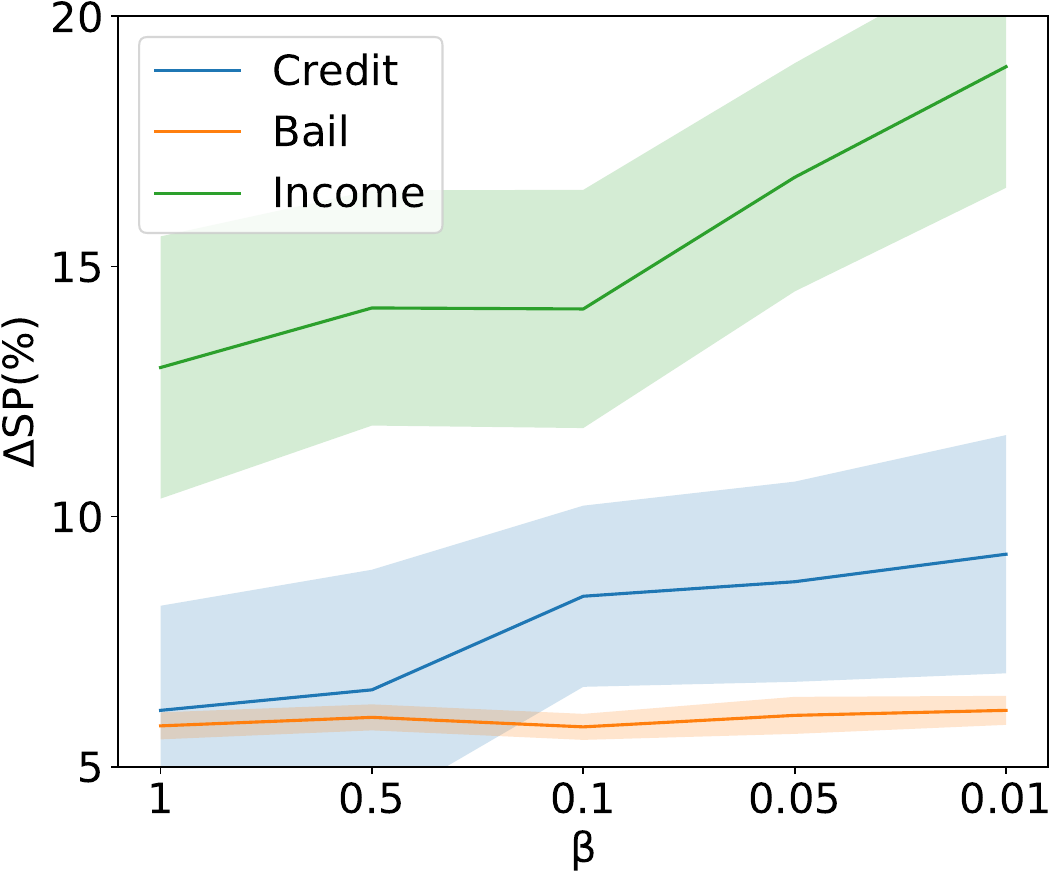}
        \label{beta_sp}
    }
    \subfigure[Impact of $\beta$ to $\Delta EO$]{
        \includegraphics[scale=0.25]{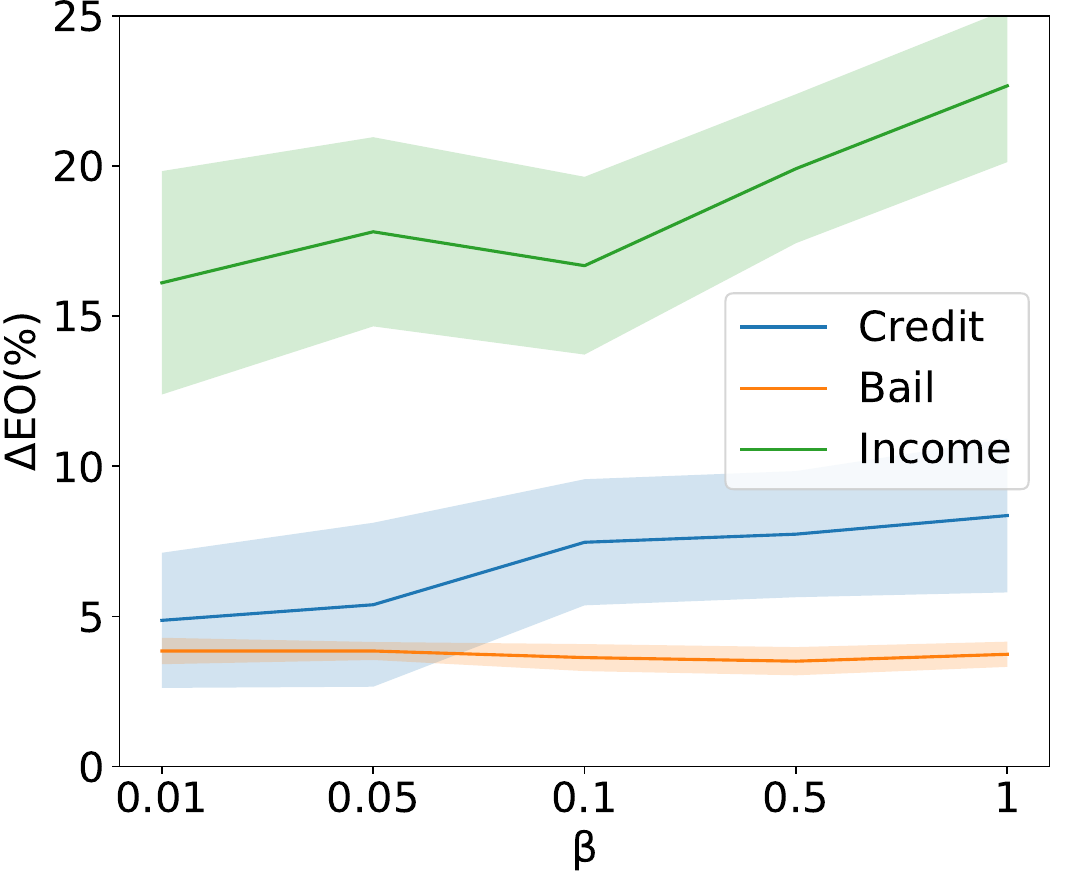}
        \label{beta_eo}
    }
    \caption{Visualization of hyperparameter $\lambda$ and $\beta$.}
    \label{vis_lamb_beta}
\end{figure*}

\subsection{Experiment Results}
In this subsection, we compare the node classification performance and fairness 
of FairMigration with the state-of-the-art models
on three different GNNs. The comparison results of AUC-ROC, $\Delta SP$, and $\Delta EO$ are
displayed in Table \ref{tab:main_comparision}.
The observations of the comparison can be summarized as:
\begin{itemize}
    \item On different GNN backbones, FairMigration achieves competitive fairness 
    with all baselines 
    while the performance of node classification is comparable. It reveals the advances of 
    FairMigration compared with other baselines.
    \item All baselines show varying degrees of unstable performance that
    sometimes perform well in fairness but badly in node classification, or on the contrary.
    For example, NIFTY shows poor performance in node classification in bail. 
    FairMigration avoids this unstable situation, achieving higher unity of 
    performance and fairness.

\end{itemize}

{
\subsection{The Improvement to Unfair Sources.}
 To investigate the impacts of the sensitive attributes' evolution,  the sensitive attributes (age) on credit are flipped when testing and the results are displayed in Figure \ref{sens_flip}. Faced with the shift of sensitive attributes, FairMigration still maintains high model utility and fairness. The biases from historical impacts and feature correlation can be alleviated by our method.
 
 To explore the influences of message propagation on fairness, we conduct the experiments of Multi-Layer Perception (MLP) on the three datasets in Table \ref{tab:message}.
It can be observed that without the message propagation, the MLP shows more fair prediction and lower model utility compared with vanilla GNNs.
With the influence of message propagation, FairMigration demonstrates competitive model utility and fairness. The biases from message passing can be alleviated by our method.
}

\subsection{Ablation Study}
In order to fully understand the contribution of each component of FairMigration, 
we conduct ablation studies, and the results are shown in Table \ref{tab:ABLATION}.
Four variants of FairMigration removing one component are defined.
The \textbf{w/o mig} denotes the variant removing the group migration (removing $L_{mig}$ in equation (\ref{pre_loss}), removing sensitive attribute value flip and setting $\lambda = 0$).
The \textbf{w/o adv} denotes the variant removing the adversarial training.
The \textbf{w/o ssf} denotes the variant removing the personalized self-supervised learning.
The \textbf{w/o wei} denotes the variant removing the reweight.
Removing group migration or personalized self-supervised learning would 
result in the most significant deterioration in fairness and little changes in node classification. { Especially, group migration improves fairness tremendously by breaking the limitation of fixed sensitive attributes.}
The combination of group migration and personalized self-supervised 
learning is a powerful technique to train fair GNNs and does not cause a significant model utility drop.
If Removing the adversarial training, FairMigration suffers from a fairness drop in most cases 
but enjoys fairness promotion in some cases. Adversarial training is an unstable strategy.
Removing the reweight, FairMigration produces slightly biased results.
The reweight technique can be a supplement to train fair GNNs.

\begin{figure*}[T]
    \centering
    \subfigure[Impact to AUC on credit]{
        \includegraphics[scale=0.25]{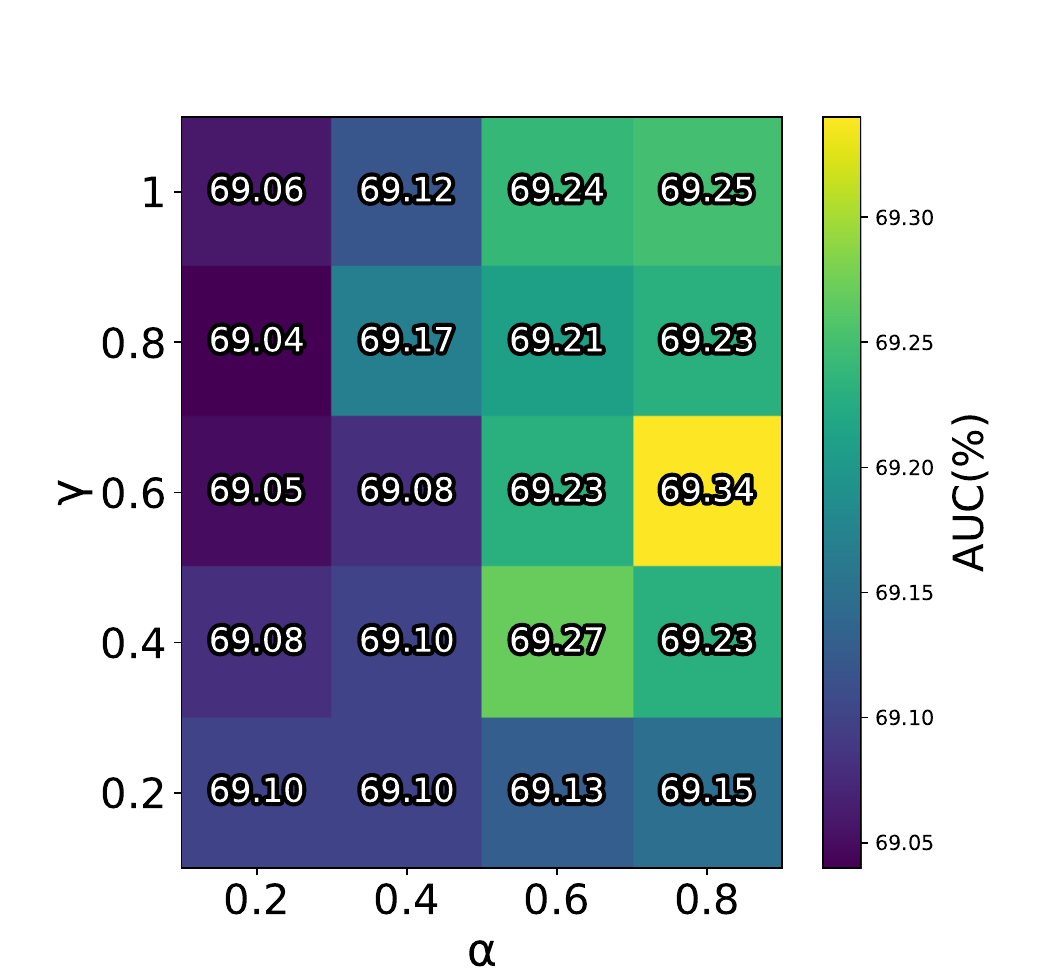}
        \label{auc_gcn_credit}
    }
    \subfigure[Impact to $\Delta SP$ on credit]{
        \includegraphics[scale=0.25]{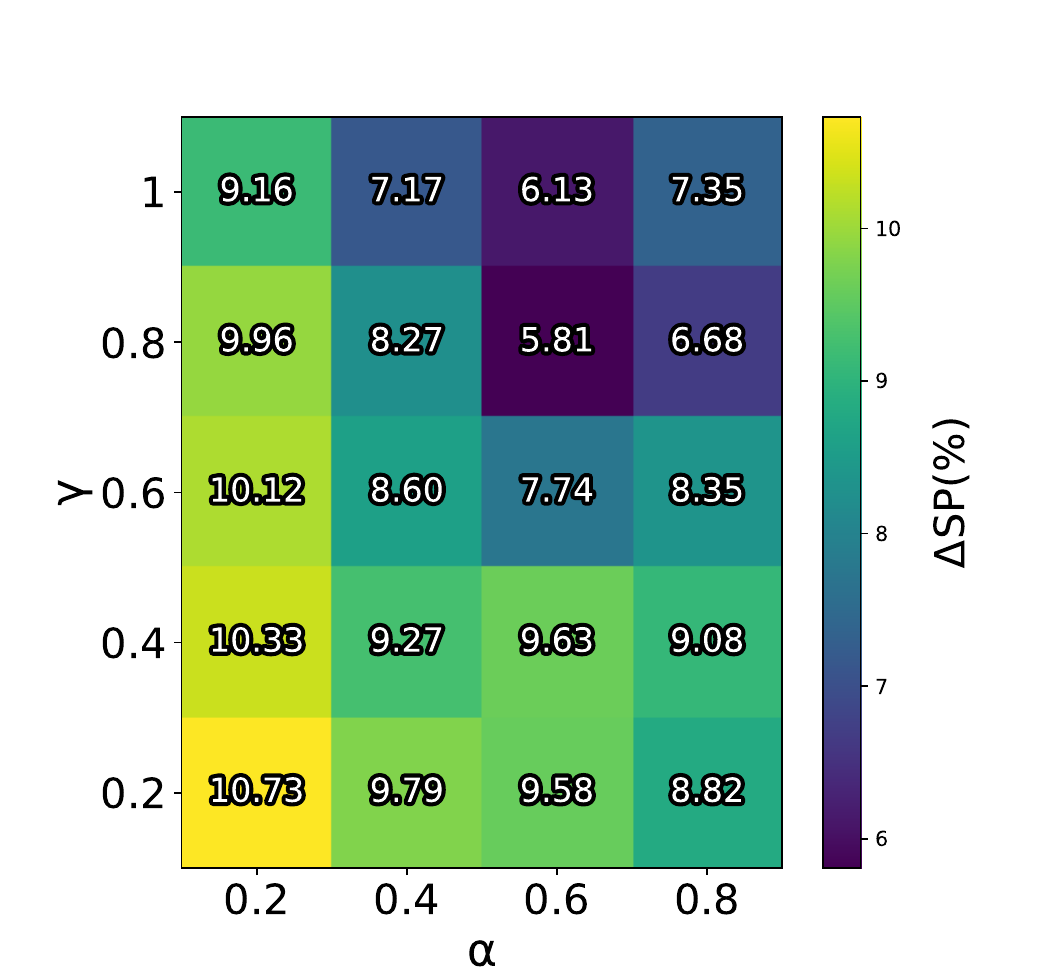}
        \label{sp_gcn_credit}
    }
    \subfigure[Impact to $\Delta EO$ on credit]{
        \includegraphics[scale=0.25]{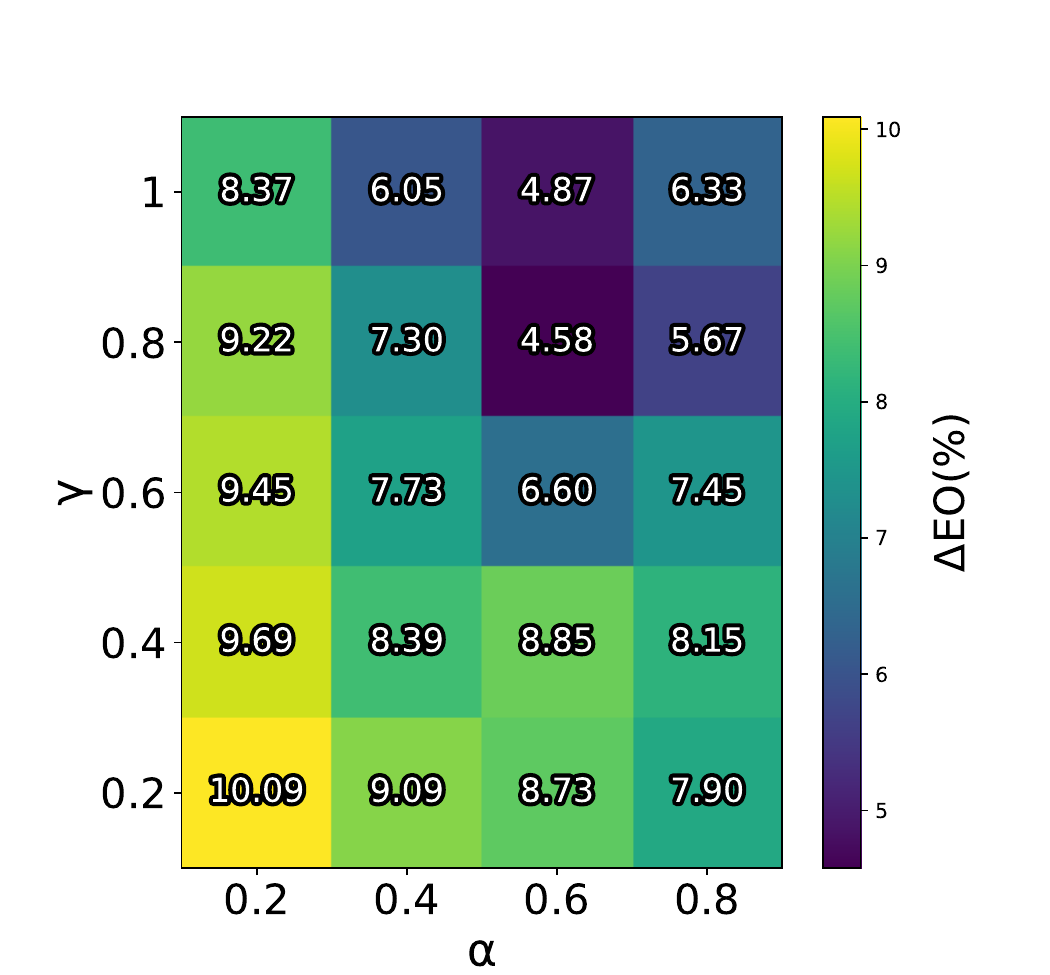}
        \label{eo_gcn_credit}
    }

    \subfigure[Impact to AUC on bail]{
        \includegraphics[scale=0.25]{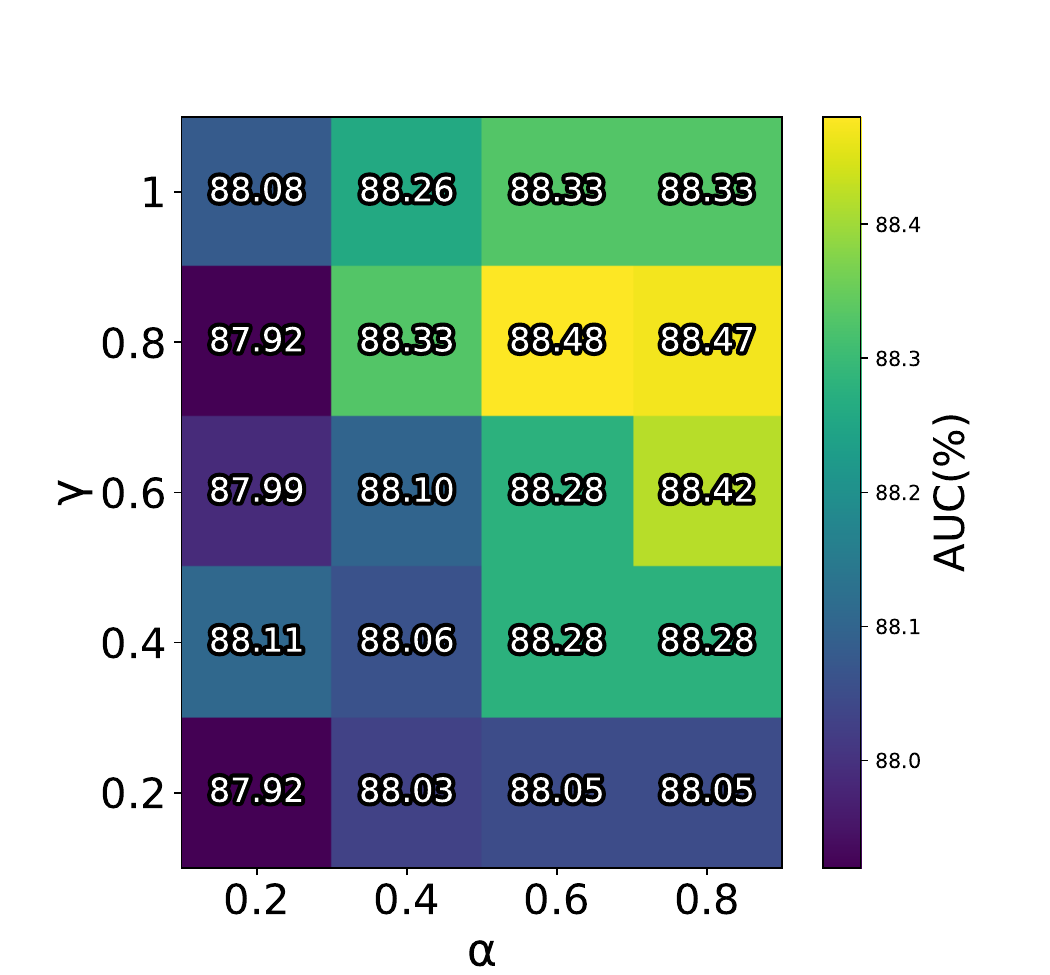}
        \label{auc_gcn_bail}
    }
    \subfigure[Impact to $\Delta SP$ on bail]{
        \includegraphics[scale=0.25]{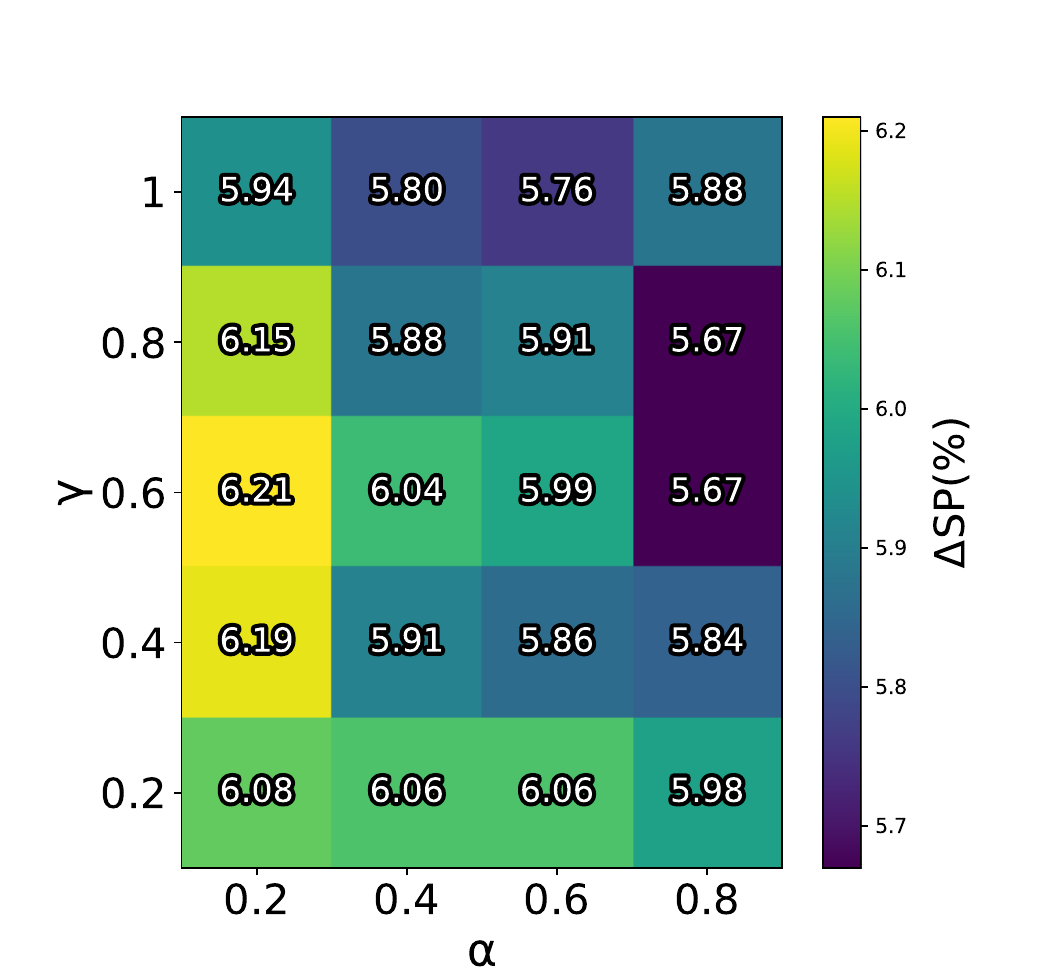}
        \label{sp_gcn_bail}
    }
    \subfigure[Impact to $\Delta EO$ on bail]{
        \includegraphics[scale=0.25]{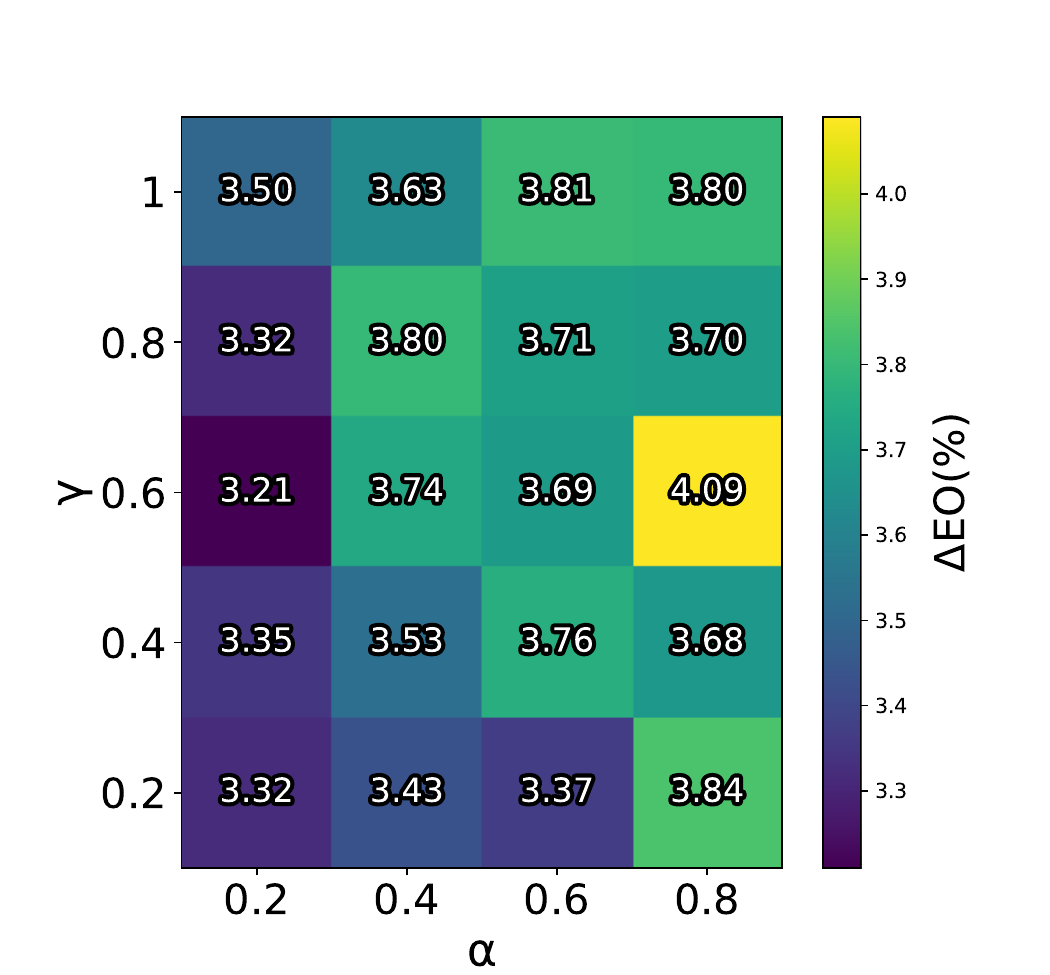}
        \label{eo_gcn_bail}
    }

    \subfigure[Impact to AUC on income]{
        \includegraphics[scale=0.25]{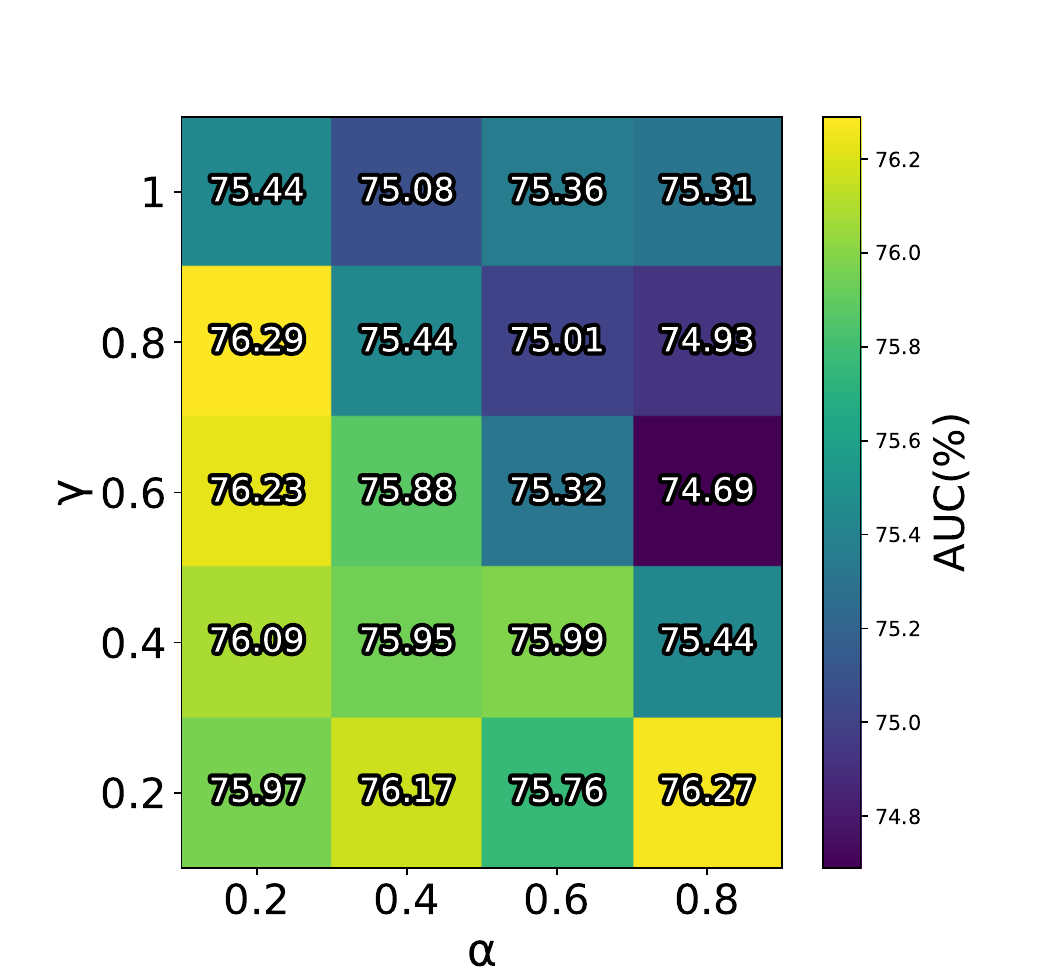}
        \label{auc_gcn_income}
    }
    \subfigure[Impact to $\Delta SP$ on income]{
        \includegraphics[scale=0.25]{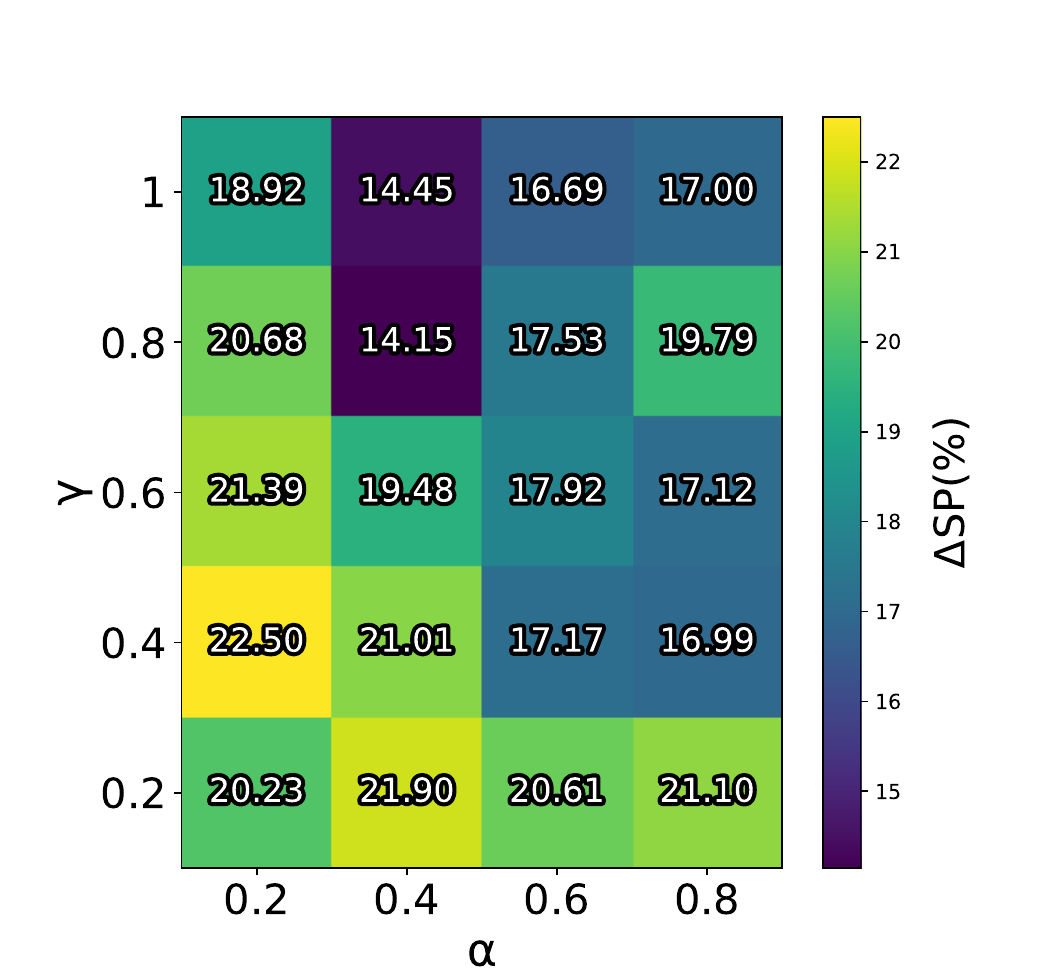}
        \label{sp_gcn_income}
    }
    \subfigure[Impact to $\Delta EO$ on income]{
        \includegraphics[scale=0.25]{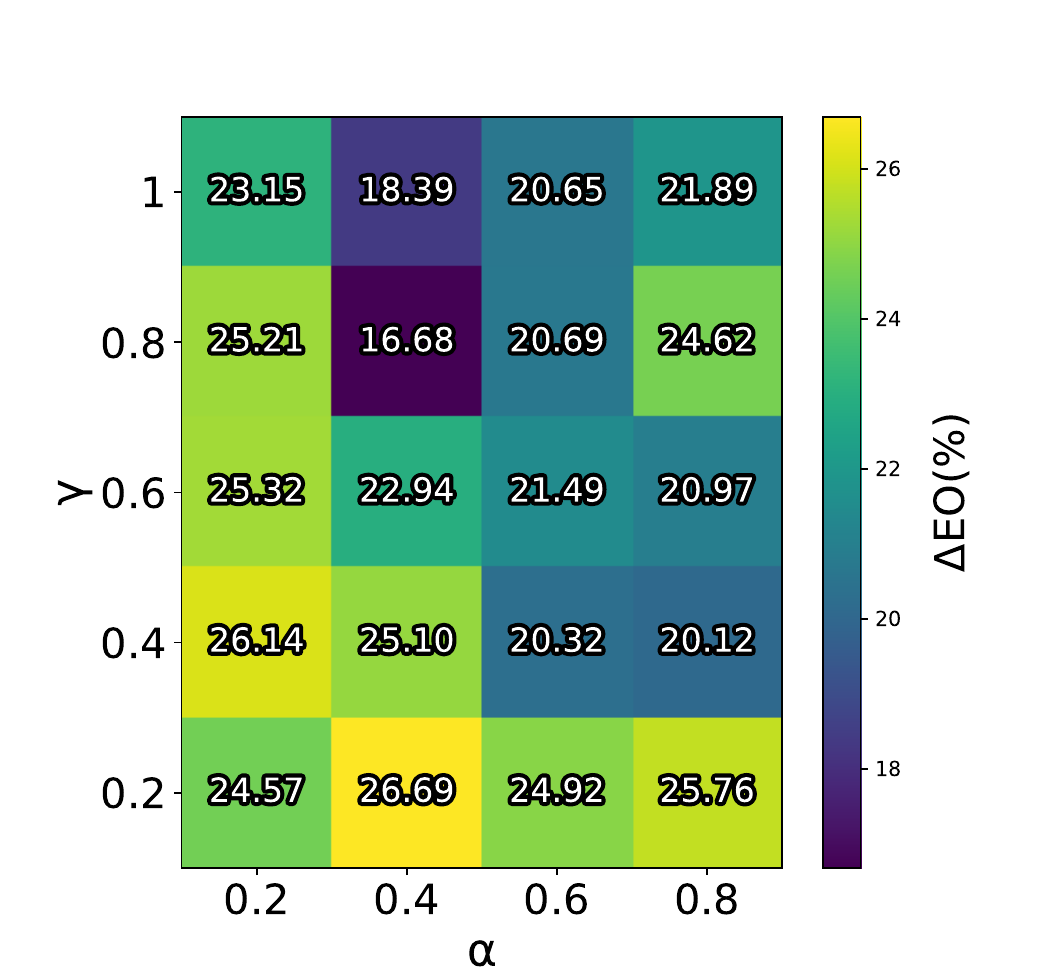}
        \label{eo_gcn_income}
    }
    
    \caption{Visualization of hyperparameter $\alpha $ and $\gamma $ of GCN.}
    \label{vis_para}
\end{figure*}

\subsection{Visualization of Group Migration}
In this section, we take GCN as the backbone as an example, 
and visualize the group migration of baselines and FairMigration during 
the supervised learning stage.
The change curve of mean and standard deviation of group similarity 
distribution is demonstrated in Figure \ref{vis_mig}.
There are the following observations:
\begin{itemize}
    \item The in-processing methods, NIFTY, FairGNN, SIND $1\%$ and FairMigration 
    gradually eliminate the similarity distribution gap between groups
    during training. The pre-posting method Edits is still unable 
    to completely get rid of the static bias information 
    even though the data is preprocessed.
    \item NIFTY and FairGNN show a certain degree of oscillation during training and
    they might not obtain a GNN fair enough with reasonable utility in some cases.
    \item SIND 1\% mainly focuses on the Wasserstein-1 distance of prediction distribution between groups and pays insufficient attention to the intra-group distance. Thus, SIND 1\% shows a low mean value and a high std value.
    \item FairMigration commendably bridges the differences 
    between the two groups in all listed cases, 
    demonstrating a remarkable ability to train fair GNNs.
\end{itemize}

\subsection{Parameter Sensitivity}
In order to investigate the impact of hyperparameters, $\lambda$, $\beta$, $\alpha$ and $\gamma$, 
we conducted experiments about parameter sensitivity, 
whose results are displayed in Figure \ref{vis_lamb_beta} and Figure \ref{vis_para}. 
The observations can be summarized as follows:
\begin{itemize}
    \item $\lambda$ is the weight of group migration constraints in supervised learning. 
    It almost doesn't impact node classification but influences fairness a lot.
    GCN achieves the highest fairness in credit, 
    bail, and income when $\lambda = 20$, $\lambda = 5$, and $\lambda = 20$, respectively.
    \item $\beta$ is the weight of adversarial training in supervised learning. 
    $\beta$ plays a little role in node classification in credit and bail 
    but disturbs the node classification in income. 
    The high value of $\beta$ would upgrade fairness in credit and income, 
    but increase little biases in bail. 
    FairMigration with GCN achieves the highest fairness in credit, 
    bail, and income when $\beta = 1$, $\beta = 0.1$, and $\beta = 1$, respectively.
    \item $\alpha$ is the trade-off between contrastive learning and reconstruction. 
    $\gamma$ is the weight of personalized self-supervised learning. 
    It is a hard task to find a combination of $\alpha$ and $\gamma$ 
    that obtain the best node classification performance and fairness simultaneously. 
    But, the optimal combination of $\alpha$ and $\gamma$ in node classification is very closed to that in fairness.
\end{itemize}

\section{Conclusion and The Future Direction}
In this paper, we explore a novel aspect of fairness, 
breaking the limitation of static sensitive attributes for training fair GNN.
We propose an innovative framework FairMigration, 
reallocating the demographic groups dynamically instead of maintaining the demographic 
group division 
following the raw sensitive attributes.
FairMigration preliminarily optimizes the encoder mainly for fairness by personalized 
self-supervised learning 
and dynamically migrates the groups. After that, 
the migrated groups and the adversarial training are adopted to constrain supervised learning.
The extensive experiments exhibit the effectiveness of FairMigration in both downstream 
tasks and fairness.
The group migration strategy is an interesting direction. FairMigration adopts a 
simple flip strategy. 
However, the optimal number of migrated groups may not be the same as the raw groups. 
Therefore, we will study the optimal migrated group division and extend it to 
multi-value sensitive attributes.

{
\section{Discussion}
This paper explores the sources of some unfairness sources and researches fairness issues. Compared with baselines, the proposed FairMigration shows superior fairness in the three datasets. However, most of the existing fairness researches are carried out on static graph-structured data. In actual application scenarios, the unbalanced quantitative relationship between intra-group edges and inter-group edges will continue to worsen over time. In addition, group representations divided by sensitive attributes are also keeping evolving. The resulting information cocoon will further isolate different groups. The impact of historical biases and time series are important factors affecting fairness. We do hope the widely applicable time series dataset with historical bias released as soon as possible, and pay more attention to this topic.
}

\section*{Acknowledgments}
The research is supported by the National Key Research and Development Program of China (2023YFB2703700), the National Natural Science Foundation of China (62176269), the Guangzhou Science and Technology Program (2023A0\newline4J0314), the National Natural Science Foundation of China and Guangdong Provincial Joint Fund (No. U1911202).
\printcredits

\bibliographystyle{cas-model2-names}

\bibliography{ref}



\end{document}